%% file: neurips_2026.tex
\documentclass{article}

\PassOptionsToPackage{numbers, compress}{natbib}

\usepackage[main, final]{neurips_2026}

\usepackage[utf8]{inputenc} 
\usepackage[T1]{fontenc}    
\usepackage{hyperref}       
\usepackage{url}           
\usepackage{booktabs}       
\usepackage{amsfonts}       
\usepackage{nicefrac}       
\usepackage{microtype}      
\usepackage{xcolor}         
\usepackage{graphicx}

\usepackage{amsmath}
\usepackage{caption}    
\usepackage{subcaption}

\usepackage{cleveref}
\usepackage{algorithm}
\usepackage{algpseudocode}
\usepackage{multirow}
\usepackage{enumitem}
\input{macros}

\usepackage{framed} % or, "mdframed"
\usepackage[framed]{ntheorem}

\usepackage[most]{tcolorbox}
\newtcolorbox{myblock}[1][]{
    colback=gray!10!white, colframe=gray!80!black,        
    fonttitle=\bfseries,        
    title=#1,                   
    boxrule=0.5mm,              
    rounded corners,            
    before skip=10pt,           
    after skip=10pt,            
}

\title{Scaling Laws for Multimodal Data Mixtures}

\author{%
  \bfseries
  Aditi Khandelwal$^{1,*,\dagger}$ \quad
  Ayush K. Tarun$^{2,*,\dagger}$ \quad
  Yixuan Xu$^{3}$ \quad
  Imanol Schlag$^{3}$ \\
  \bfseries
  Golnoosh Farnadi$^{1}$ \quad
  Siva Reddy$^{1}$ \quad
  Luke Zettlemoyer$^{4}$ \quad
  Antoine Bosselut$^{2}$ \\[6pt]
  \normalfont
  $^{1}$MILA - Quebec AI Institute, McGill University
  \quad
  $^{2}$EPFL \quad
  $^{3}$ETH \quad
  $^{4}$University of Washington
}

\newcommand{\Neff}{N_{\mathrm{eff}}}
\newcommand{\Deff}{D_{\mathrm{eff}}}

\begin{document}

% \maketitle
\maketitle

\begingroup
\def\thefootnote{}\footnotetext{%
  $^{*}$Equal contribution. \quad
  $^{\dagger}$Corresponding author (\texttt{aditi.khandelwal@mila.quebec, ayush.tarun@epfl.ch}).%
}
\endgroup

\input{sections/abstract}
\input{sections/introduction}

\input{sections/method}

\input{sections/experiments}
\input{sections/unimodalscalinglaws}
\input{sections/results}
\input{sections/validation}

% \input{sections/ablations}
\input{sections/related_works}
\input{sections/conclusion}
\input{sections/limitation_and_future_work}
\input{sections/impact_statement}
\input{sections/acknowledgments}

\bibliography{custom}
\bibliographystyle{plainnat}

\input{sections/appendix}

% \newpage
% \clearpage
% \input{checklist}

\end{document}

%% file: macros.tex
\usepackage{todonotes}
\usepackage[normalem]{ulem}

\makeatletter
\newcommand*\iftodonotes{\if@todonotes@disabled\expandafter\@secondoftwo\else\expandafter\@firstoftwo\fi}  % defines \iftodonotes{<true>}{<false>}, thanks to https://tex.stackexchange.com/questions/126559/conditional-based-on-packageoption
\makeatother

\newcommand{\rev}[1]{{\color{red}#1}}

%% file: sections/abstract.tex
\begin{abstract}

Frontier AI systems are increasingly natively multimodal, jointly pretrained on multiple modalities, such as text, vision, and audio. However, the scientific understanding of the multimodal data mixtures used to pretrain these systems --- how much of each modality to mix, and at what potential interference cost to the others --- is largely absent or gatekept. In this work, we derive data mixing scaling laws for three modalities: text, vision, and audio. Multimodal Large Language Models (MLLMs) increasingly adopt Mixture-of-Experts (MoE) architectures, as MoEs scale efficiently and partially limit cross-modal interference. We therefore use MoEs to conduct 268 experiments ranging from 457M to 8.3B parameters, on up to 150 billion tokens. Our work leads to multiple actionable insights into multimodal data mixing for training MLLMs. First, compute optima are modality-specific: audio modality is model-heavy, favoring scaling parameters over tokens, while text and vision require balanced scaling of model size and data. Second, repeating vision or audio data beyond 2x yields negligible benefits. Finally, cross-modal interactions are highly asymmetric: vision strongly interferes with audio, {whereas audio's interference with learning text and vision is negligible.} Overall, we provide a principled foundation for understanding the multimodal data mixtures needed to train frontier MLLMs.

\end{abstract}

% \vspace{-1em}

%% file: sections/introduction.tex
\section{Introduction}

% \section{Introduction}

Frontier AI systems are increasingly natively multimodal
\citep{google_gemini3_2025, team2026qwen3, openai2024gpt4ocard, meta2025llama4}. Yet details of how these systems are trained remain largely opaque, kept behind closed doors as trade secrets. Public scaling recipes \citep{kaplan2020scalinglawsneurallanguage, hoffmann2022trainingcomputeoptimallargelanguage} are therefore often insufficient for practitioners training multimodal systems, where pretraining introduces additional design decisions: data mixtures, architectural choices, training strategies, and modality-specific algorithms. One trend that has stood out is the convergence toward \textbf{Mixture-of-Experts (MoE)} models for multimodality \citep{comanici2025gemini25pushingfrontier, meta2025llama4, google_gemini3_2025, team2026qwen3, openai2024gpt4ocard}. This convergence is natural for two reasons. First, MoEs scale capacity through sparse activation \citep{zoph2022stmoedesigningstabletransferable}, decoupling parameter count from compute and making it tractable to grow models large enough to serve heterogeneous signals. Second, sparse routing permits modality-specific expert allocation, allowing parameters to specialize to distinct modalities and partially insulate them from one another \citep{tong2026beyond, lin2024momaefficientearlyfusionpretraining, li2024unimoescalingunifiedmultimodal}. Despite this architectural advantage, jointly training text, vision, and audio still introduces non-trivial optimization complexity \citep{ramakrishnan2018overcoming, gat2020removing}, as modalities have distinct feature representations, follow heterogeneous training dynamics \citep{liang2024mixture}, and compete for the same finite parameter capacity and routing budget, often interfering with one another.

Data imbalance further compounds this capacity competition. In realistic multimodal pretraining mixtures, modalities differ substantially in dataset availability, filtering quality, and target mixture proportions \citep{zou2026caption, tseng2025revisiting, tian2025navil}. As a result, non-text modalities such as vision and audio are often trained in data-constrained regimes, where maintaining a desired mixture may require repeated passes over the same examples. While the repetition tolerance of text is well characterized \citep{muennighoff2025scalingdataconstrainedlanguagemodels}, the saturation point for repeated vision or audio data remains unknown. Prior work \citep{aghajanyan2023scaling, sun2024scalinglawhypothesismultimodal, shukor2025scaling, manakul2026scaling, tian2025navil} has begun to address parts of this picture, including bimodal text--vision interactions and individual continuous-modality scaling laws. However, a \emph{unified} framework that jointly handles text, vision, and audio under realistic, repetition-aware, data-imbalanced mixtures is still missing.

In this work, we introduce a unified scaling-law {framework} for native multimodal MoE pretraining that models a given target modality's validation loss in the presence of other modalities. Starting from the standard Chinchilla form \cite{hoffmann2022trainingcomputeoptimallargelanguage}, our formulation accounts for three effects that arise in realistic multimodal mixtures. First, we model \emph{directional cross-modal transfer}, capturing when a source modality helps or hurts a target modality. Second, we model \emph{mixture-dependent capacity sharing}, capturing the loss of effective capacity when modalities compete for the same finite parameter and routing budget in MoE architectures. Third, we model \emph{repetition saturation}, capturing the diminishing utility of repeated passes over scarce non-text data. The resulting law reduces to the standard Chinchilla form in the unimodal limit and generalizes prior bimodal interaction laws \citep{aghajanyan2023scaling} to directional, unbalanced, and repetition-aware mixtures. We fit the law on a controlled suite of MoE pretraining runs spanning 457M to 8.3B total parameters and up to 150B tokens per run, totaling roughly 300k GH200-hours.  Collectively, our experiments yield several insights for native multimodal pretraining:

\begin{figure}[t] 
    \centering
    \includegraphics[width=\textwidth]{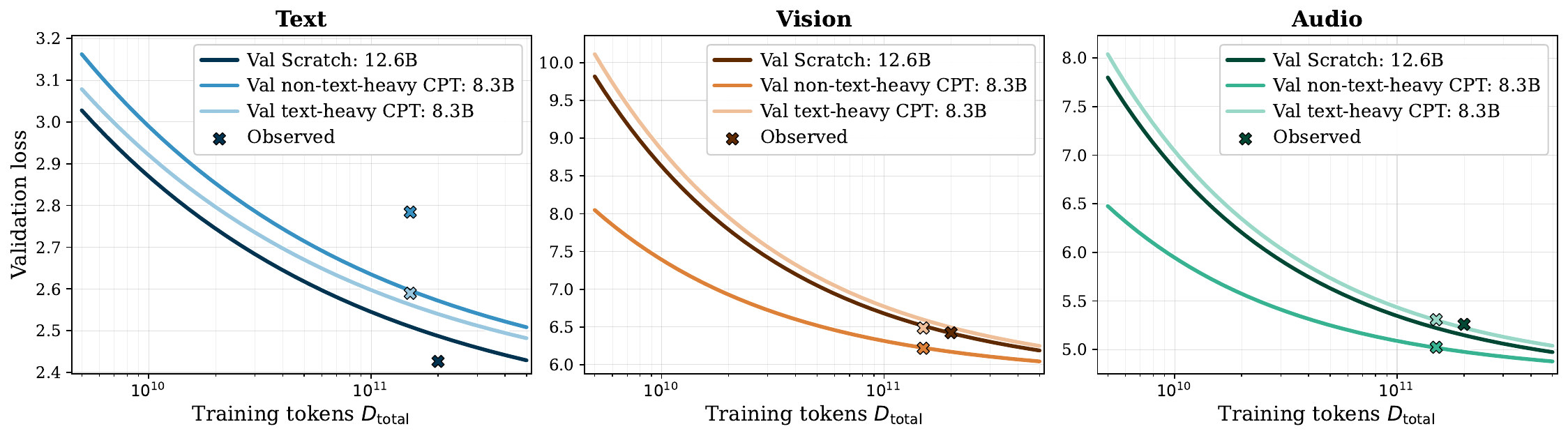}
    \caption{Predicted loss curves and observed values for OOD training scales and regimes: (i) \textit{Val Scratch}: a 12.6B-A3.8B MoE trained for 200B tokens (185B text, 7.5B vision, 7.5B audio); (ii) \textit{Val non-text-heavy CPT}: continual pretraining (CPT) of a text-only 8.3B-A2.6B MoE on 50B tokens (10B text, 20B (unique 10B$\times$2) each vision and audio); and (iii) \textit{Val text-heavy CPT}: CPT of text-only 8.3B-A2.6B MoE on 40B text, 5B each of vision and audio. Observed points are very close to the predicted curves. The only exception is the text loss in (ii), where CPT causes a significant spike in text loss (Appendix \ref{app:validation_at_scale}) and does not stabilize, unlike (iii), which has a higher text proportion.}
    % We hypothesize that longer training would stabilize the text loss and bring it closer to our law’s prediction.}
    \label{fig:sample_interaction}
\end{figure}

\begin{itemize}[leftmargin=1.2em, itemsep=1pt, topsep=2pt]
    \item \textbf{Compute allocation is modality-specific.}
    Text and vision should be scaled by growing parameters and tokens in tandem. Audio follows a more model-heavy regime, where extra compute is better spent on increasing capacity than adding more audio tokens. 
    \item \textbf{Cross-modal transfer is asymmetric.}
    {Text and vision interfere with one another, whereas audio's interference with both modalities is negligible, making it an effectively inert addition to the data mixture.} Vision is the most disruptive source, especially for audio, making it the most expensive modality to mix in when audio is also an evaluation target.
    \item \textbf{Repeated non-text data saturates quickly.}
    The repetition tolerance observed for text \citep{muennighoff2025scalingdataconstrainedlanguagemodels} does not directly carry over to continuous modalities: vision and audio both show sharply diminishing returns after roughly two passes over the same data. Beyond this point, compute is better spent on additional unique data or larger models.
\end{itemize}

Finally, we show that the fitted law extrapolates beyond its training distribution (Figure \ref{fig:sample_interaction}) to larger compute budgets, out-of-distribution (OOD) mixture compositions, and continual pretraining. Across three held-out runs, 8 of 9 modality loss predictions match observed validation loss within 2.6\%, and 6 of 9 within 1.1\% with a correlation $>$ 0.9.

%% file: sections/method.tex
\section{Method}
\label{sec:method}

To systematically study how modalities interact during their joint optimization, we introduce a unified multimodal scaling framework for text, vision, and audio. Standard neural scaling laws model the validation loss of a target modality as a predictable power-law function of model capacity and token count \citep{kaplan2020scalinglawsneurallanguage, hoffmann2022trainingcomputeoptimallargelanguage} as follows:
\begin{equation}
    \label{chinchilla-basic}
    L_m(N,D_m)
    =
    E_m
    +
    \frac{A_m}{N^{\alpha_m}}
    +
    \frac{B_m}{D_m^{\beta_m}}
\end{equation}

Here, 
\(m\in\mathcal{M}\), where \(\mathcal{M}=\{\text{text},\text{vision},\text{audio}\}\). Let \(N\) denote the total parameter count of the
model (including non-active parameters and embeddings \cite{pearce2024reconciling}), \(D_m\) the number of sampled training tokens from modality \(m\).  The first term \(E_m\) captures
the irreducible loss for modality \(m\), while the second and third terms
capture finite-capacity and finite-data effects, respectively
\citep{kaplan2020scalinglawsneurallanguage,hoffmann2022trainingcomputeoptimallargelanguage}.

We extend this formulation to capture three factors: (1) data-based positive transfer or interference between modalities, (2) parametric capacity competition between modalities, and (3) the penalty incurred by oversampling scarce modality data. In this regard, we propose replacing $N$ and $D_m$ with $\Neff^m=\Neff(N, m, \mathcal{\hat{M}})$, and $\Deff^m=\Deff(D_m,\mathcal{\hat{M}})$ which denote the effective number of parameters and tokens, also taking into account the multimodal data mixture $\mathcal{\hat{M}}$. Intuitively, $\Neff$ measures how much of the MoE's parameters are effectively available to a target modality once it must be shared with other modalities, while $\Deff$ measures how many of that modality's tokens are effectively informative once we discount for the diminishing returns of data repetition and credit (or debit) it for cross-modal transfer from the co-trained modalities. Concretely, we make \emph{one} modification to the $N^{\alpha_m}$ term (Section~\ref{ssec:capacity}) and \emph{two} modifications to the $D_m^{\beta_m}$ term (Section~\ref{ssec:data}, covering data repetition and cross-modal transfer). The following sections describe the construction of both of these.

\begin{figure}[t]
    \centering
    \includegraphics[width=0.9\textwidth]{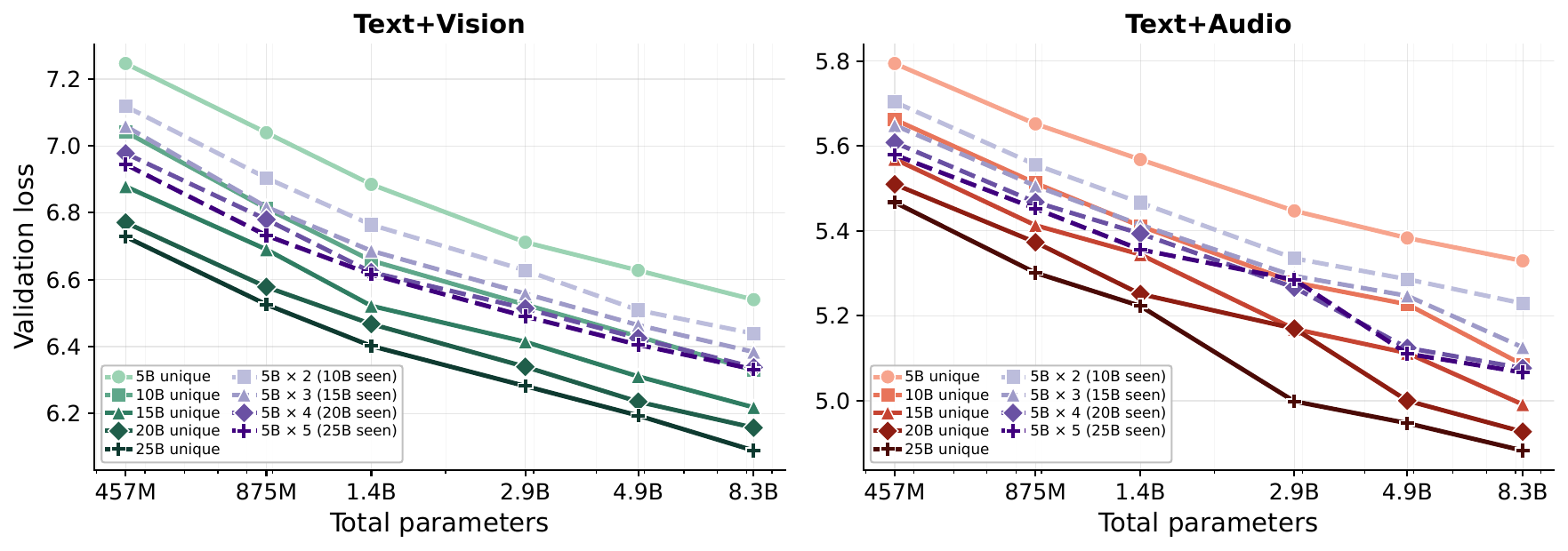}
    \caption{Validation loss as a function of model size for text+vision (left) and text+audio (right) training at $100$B total tokens, averaged across modality-specific evaluation sets. Solid curves correspond to training with $5$B--$25$B \emph{unique} non-text tokens; dashed purple curves repeat $5$B unique tokens $2$--$5$ times. At every model scale, repeating data 5$\times$ underperforms training on roughly $15$B unique tokens, motivating the diminishing-returns form of $\phi_m(r_m)$ in Equation~\ref{eq:repetition_phi}.}
    \label{fig:vision_audio_rep}
\end{figure}

\subsection{Mixture-dependent Capacity Sharing $\Neff^{(m)}$}
\label{ssec:capacity}

Multimodal training changes how a finite model
allocates its representational and optimization capacity: non-target modalities consume optimizer steps, gradients,
routing bandwidth, and shared representation capacity. This forced sharing can hurt the target
modality even when the source modality contains useful information \cite{aghajanyan2023scaling}, a phenomenon
analogous to capacity competition in multi-task learning \cite{fifty2021efficiently, standley2020tasks}. To capture the cost of sharing finite MoE capacity, we replace $N$ with a target-specific effective capacity parameterized by the \textbf{capacity-sharing coefficient} $c_m$:
\begin{equation} \label{eq:capacity}
    N_{\mathrm{eff}}^{(m)}
    =
    N
    \exp\!\left(
        c_m(1-h_m)
    \right).
\end{equation}
Here, $h_m = {D_m}/{\sum_{\ell\in\mathcal{M}} D_\ell}$ is the mixture weight of modality $m$ and therefore, \(1-h_m=\sum_{k\neq m} h_k\) denotes the total non-target mixture mass. The
coefficient \(c_m\) is a signed log-capacity coefficient. If \(c_m<0\), then
non-target data reduces the effective capacity available to modality \(m\) {(we term this \textit{capacity-dilution})};
if \(c_m>0\), non-target data has a beneficial effect based on shared capacity.

% When
% we want to impose pure dilution, we write \(c_m=-\gamma_m\) with
% \(\gamma_m\ge 0\), giving
% \[
%     N_{\mathrm{eff}}^{(m)}
%     =
%     N\exp[-\gamma_m(1-h_m)].
% \]

The exponential form follows from a simple constant fractional change assumption on
the log-capacity scale: ${\partial \log N_{\mathrm{eff}}^{(m)}}/{\partial (1-h_m)} = c_m$. Thus each additional unit of non-target mixture mass changes the effective
target capacity by a fixed multiplicative factor. It also ensures
\(N_{\mathrm{eff}}^{(m)}>0\) and yields a simple first-order interaction term.
Indeed,
\begin{equation}
    \frac{A_m}{(N_{\mathrm{eff}}^{(m)})^{\alpha_m}}
    =
    \frac{A_m}{N^{\alpha_m}}
    \exp[-\alpha_m c_m(1-h_m)]
\enskip \approx \enskip
    \frac{A_m}{N^{\alpha_m}}
    \left[
        1-\alpha_m c_m(1-h_m)
    \right],
\end{equation}
    
where the approximation holds when $|c_m|$ is small. If \(c_m<0\), this adds a positive penalty
\(\alpha_m |c_m|(1-h_m)A_m/N^{\alpha_m}\), \textbf{which vanishes as model capacity
increases.} This is consistent with the intuition that increasing model capacity can attenuate negative transfer because of capacity \cite{aghajanyan2023scaling}.

We use the scalar \(c_m(1-h_m)\), instead of the directional capacity
matrix \(\sum_{k\neq m}u_{k\to m}h_k\), as the formulation with bidirectional capacity overfits on our limited-scale data (See Appendix \ref{app:validation} for comparison).

\subsection{Effective Target-Modality Data $\Deff^{(m)}$}
\label{ssec:data}

We hypothesize two factors that move $\Deff^{(m)}$ away from the raw unique-token count $D_{\mathrm{unq}}^{(m)}$: a within-modality \emph{repetition} factor $\phi_m(r_m)$ capturing the diminishing returns of re-passing the same examples, and a between-modality \emph{transfer} factor $\exp\!\big(\sum_{k\neq m} v_{k\to m}\, h_k\big)$ capturing help (or interference) from co-trained modalities.

\paragraph{Repetition-aware target data.}\label{ssec:repetition}
In real-world mixtures, modalities like audio are seen multiple times while text is seen only once. To model the diminishing returns of data repetition, let \(D_{\mathrm{unq}}^{(m)}\) be the number of unique tokens from modality
\(m\) and  $r_m={D_m}/{D_{\mathrm{unq}}^{(m)}} \ge 1$ be the repetition ratio. We define the repetition-aware effective target-modality data,  $D_m^r = D_{\mathrm{unq}}^{(m)} \, \phi_m(r_m)$, where \(\phi_m(r_m)\) is a return-on-repetition function. \(\phi_m(r_m)\) should ideally encode the following: the first pass over unique data is fully valuable (\(\phi_m(1)=1\)), repeated passes still
help (\(\phi_m'(r)>0\)), but the marginal value of each additional repeat decays (\(\phi_m''(r)<0\)). Our repetition ablations (Figure~\ref{fig:vision_audio_rep}) also show that repeated
non-text data quickly falls behind fresh data, motivating an exponentially decaying marginal-utility
model for $\phi_m$. We therefore use the following formulation, inspired by similar prior formulations \cite{muennighoff2025scalingdataconstrainedlanguagemodels, longpre2025atlas}:

\begin{equation} \label{eq:repetition_phi}
    \phi_m(r_m)
    =
    1
    +
    R_m^\star
    \left(
        1-\exp\!\left(-\frac{r_m-1}{R_m^\star}\right)
    \right),
\end{equation}
where \(R_m^\star>0\) is a modality-specific repetition scale. This form would imply that as \(r_m\to\infty\), $\phi_m(r_m)\to 1+R_m^\star$. Thus, \(R_m^\star\) can be interpreted as the maximum number of additional fresh-data-equivalent epochs available through repetition.

\paragraph{Directional cross-modal transfer.}\label{ssec:transfer}
Next, we model how the presence of a source modality $k$'s data alters the learning efficiency of a target modality $m$.  A natural starting point is an additive effective-data model (assuming no repetition) \cite{longpre2025atlas},
\begin{equation}
    D_m^t
    =
    D_m
    +
    \sum_{k\neq m}
    \tau_{k\to m}D_k,
\end{equation}

where \(\tau_{k\to m}>0\) indicates positive transfer and
\(\tau_{k\to m}<0\) indicates interference. However, for negative
\(\tau_{k\to m}\), additive effective data can become non-positive in highly
skewed mixtures ($|D_k|\gg|D_m|$), making the power-law term ill-defined. This is especially
problematic when a scarce target modality is trained with a large amount of non-target modality data.

To preserve the interpretation of transfer while guaranteeing positivity, we
model transfer on the log-effective-data scale:
\begin{equation}
    \log D_m^{t}
    =
    \log D_m
    +
    \sum_{k\neq m}
    v_{k\to m} h_k.
\quad \Rightarrow \quad
    D_m^t
    =
    D_m
    \exp\!\left(
        \sum_{k\neq m}
        v_{k\to m} h_k
    \right).
    \label{eq:log_transfer}
\end{equation}

Here, the variable \(v_{k\to m}\) is the \textbf{directional transfer coefficient}. Using an exponential function ensures that our effective data count always remains strictly positive, even in the event of severe cross-modal interference (a highly negative \(v_{k\to m}\)) or a highly skewed data mixture. Positive
\(v_{k\to m}\) means that increasing the mixture weight of modality \(k\) increases the effective data for target modality \(m\); negative \(v_{k\to m}\) means that modality \(k\) interferes with the learning of modality \(m\).

This exponential form can also be viewed as a locally additive transfer model. Let $g_m = \sum_{k\neq m} v_{k\to m}h_k$, then for small transfer magnitudes, a first-order Taylor expansion gives:
\begin{equation}
    D_m^t
    =
    D_m \exp(g_m)
    \approx
    D_m
    \left(
        1 + g_m + O(g_m^2)
    \right)
    =
    D_m
    +
    \sum_{k\neq m}
    v_{k\to m} h_k D_m
    +
    O(g_m^2)D_m.
\end{equation}
    
Thus, to first order, the log-linear model behaves like an additive
effective-data model with local additive coefficient $\tau_{k\to m}^{\mathrm{local}} = v_{k\to m} h_k {D_m}/{D_k}$. When data are unrepeated, and total tokens are fixed,
\(D_k/D_m = h_k/h_m\), so
\(\tau_{k\to m}^{\mathrm{local}} = v_{k\to m}h_m\). Hence the proposed
parameterization is a first-order log-linear analogue of additive transfer,
but it remains positive for all finite \(v_{k\to m}\) and all mixture weights. Substituting Equation~\ref{eq:log_transfer} into the data term gives
\begin{equation}
    \frac{B_m}{(D_m^{t})^{\beta_m}}
    =
    \frac{B_m}{D_m^{\beta_m}}
    \exp\!\left(
        -\beta_m
        \sum_{k\neq m} v_{k\to m}h_k
    \right)
    \enskip\approx \enskip
    \frac{B_m}{D_m^{\beta_m}}
    \left[
        1
        -
        \beta_m
        \sum_{k\neq m} v_{k\to m}h_k
    \right],
\end{equation}
    
where the approximation stands for small $|v_{k\to m}|$. Therefore, positive transfer reduces the finite-data loss term, while interference increases it.

Combining the repetition correction $\phi_m(r_m)$ from the previous paragraph with this directional transfer factor gives the full effective target-modality data,
\begin{equation}
    \Deff^{(m)}
    =
    D_{\mathrm{unq}}^{(m)} \, \phi_m(r_m) \,
    \exp\!\left(
        \sum_{k\neq m} v_{k\to m} h_k
    \right),
    \label{eq:deff}
\end{equation}
the data-side analogue of the capacity-sharing $\Neff^{(m)}$ from Section~\ref{ssec:capacity}.

\subsection{Unified Multimodal Scaling Law}
\label{ssec:unified}

Plugging $\Neff^{(m)}$ (Section~\ref{ssec:capacity}) and $\Deff^{(m)}$ (Section~\ref{ssec:data}) into Equation~\ref{chinchilla-basic} yields the final law:
\begin{equation}
    L_m
    =
    E_m
    +
    \frac{A_m}
    {
        \left[
            N\exp(c_m(1-h_m))
        \right]^{\alpha_m}
    }
    +
    \frac{B_m}
    {
        \left[
            D_{\mathrm{unq}}^{(m)}
            \phi_m(r_m)
            \exp\!\left(
                \sum_{k\neq m}
                v_{k\to m}h_k
            \right)
        \right]^{\beta_m}
    }.
    \label{eq:unified_law}
\end{equation}
This equation reduces to the standard unimodal law when \(h_m=1\),
\(h_k=0\) for \(k\neq m\), and \(r_m=1\). Our unified formulation also generalizes prior bimodal interaction laws \cite{aghajanyan2023scaling} to directional, unbalanced, and repetition-aware mixtures. In Appendix~\ref{app:proof_bimodal}, we elaborate on how our log-linear capacity and transfer terms recover existing additive bimodal formulations under small-interaction approximations. We also ablate scaling law formulations without each of our proposed components and show that our formulation generalizes the best. The analysis can be found in Appendix \ref{app:validation}.

%% file: sections/experiments.tex
\section{Experimental Setup}
\label{sec:experiments} 
To empirically fit the variables in our unified scaling framework, we design a highly controlled suite of pretraining experiments across six Mixture-of-Experts (MoE) parameter scales (457M to 8.3B total parameters) on curated multimodal mixtures.

\textbf{Models.}
\citet{wan2026holistic} show in MoEs, the hidden dimension ($d_{\text{model}}$) is rank-preserving, meaning the relative performance ordering of models remains stable across compute and parameter scales for different $d_{\text{model}}$ values. This allows us to vary model capacity via the hidden dimension and attribute downstream performance changes to multimodal data mixtures rather than architectural confounds. All models in our experiments utilize 32 layers, Grouped Query Attention \citep{ainslie2023gqa} with 16 attention heads and $4$ key-value heads, $1$ shared expert, $16$ routable experts (of which $2$ are chosen for each token), an expert hidden dimension of $d_{\text{model}}$, and untied embeddings. $d_{model}$ varies from 384 to 2048, and SwiGLU activation is used along with QK-Norm for training stability \citep{chameleonteam2025chameleonmixedmodalearlyfusionfoundation}. More training details and exact model sizes can be found in Appendix \ref{app:train_config}.

\textbf{Tokenizers.} For text, we use the tokenizer from Mistral-Nemo \citep{mistralnemo} (vocabulary size $131$K), Emu3.5 tokenizer \citep{cui2025emu35nativemultimodalmodels} for images (vocabulary size $131$K)  and WavTokenizer (vocabulary size $4$K) \citep{ji2024wavtokenizer} with a token rate of $40$ tokens/second for audio.

\textbf{Datasets and Evaluations.} We use high-quality, permissive \footnote{CC BY 4.0 and similarly licensed} datasets for each modality to maximize signal in our limited-scale experiments and ensure training stability. Our data composition deliberately mimics the distribution mixtures used in real-world MLLM pretraining, following the methodologies of Apertus for text \cite{hernandez2025apertus}, Chameleon for vision \cite{chameleonteam2025chameleonmixedmodalearlyfusionfoundation}, and Voxtral for audio \cite{liu2025voxtral}. Briefly, the text corpus balances general knowledge, reasoning, and coding; the vision corpus employs both standard and reversed image-caption pairs to encourage bidirectional alignment; and the audio corpus mixes speech-only datasets with synthetic interleaved audio-text sequences. Our validation sets consist of 100 million token subsets drawn from these same distributions. Moreover, note that the vision and audio losses include the loss on text tokens from their respective validation sets. Full dataset breakdowns, sourcing details, and license filtering protocols are provided in Appendix~\ref{app:datasets}.

\textbf{Hyperparameters.} \label{small_sec: hyperparams} Following the Step Law \citep{li2025predictablescaleistep}, we perform a hyperparameter sweep over learning rates and batch sizes (detailed in Appendix~\ref{app:lrbs}). Across all model scales and modalities, we find that applying a \textbf{1$\times$ Step Law learning rate and a 2$\times$ Step Law batch size} consistently achieves the lowest or close to the lowest loss, and we use this configuration for all subsequent experiments. To check seed sensitivity, we ran the smallest and largest models for $30$B tokens with three different seeds and observed $<0.3\%$ variance in validation loss; we therefore conduct the rest of our experiments with a single seed.

\textbf{Experiment Grid}.
For each model scale, we run a controlled grid of pretraining experiments designed to isolate the main factors in our scaling law: unimodal scaling, repetition saturation, pairwise cross-modal transfer, and three-way modality interference. All runs are trained in an autoregressive fashion and evaluated on the held-out modality-specific validation sets described above, and the resulting losses are used to jointly fit the scaling law in Section~\ref{ssec:unified}. Our grid consists of the following: %\aditi{add causal language modeling objective}

\begin{itemize}[leftmargin=1.2em,topsep=0pt,itemsep=2pt,parsep=0pt,partopsep=0pt]
    \item \textbf{Unimodal scaling.} (1) Text-only baselines: $10$B, $20$B, $40$B, and $100$B tokens; (2) vision and audio baselines: $5$B, $10$B, $15$B, $20$B, and $25$B tokens.
    \item \textbf{Text--vision and text--audio mixtures.} Text-skewed: (1) $D_{\text{text}}+D_{\text{vision}}=100$B with $D_{\text{vision}} \in \{5,10,15,20,25\}$B, and (2) $D_{\text{text}}+D_{\text{audio}}=100$B with $D_{\text{audio}} \in \{5,10,15,20,25\}$B. Balanced: (3) $25$B$+25$B text--vision and (4) $25$B$+25$B text--audio.
    
    \item \textbf{Repetition ablations.} (1) $D_{\text{text}}+D_{\text{vision}}=100$B with $D_{\text{unq}}^{\text{vision}}=5$B and $r_m \in \{1,2,3,4,5\}$; (2) $D_{\text{text}}+D_{\text{audio}}=100$B with $D_{\text{unq}}^{\text{audio}}=5$B and $r_m \in \{1,2,3,4,5\}$.
    
    \item \textbf{Vision--audio and three-way mixtures.} (1) $25$B vision $+$ $25$B audio; (2) $D_{\text{text}}+D_{\text{vision}}+D_{\text{audio}}=150$B with $D_{\text{text}}=100$B and $(D_{\text{vision}}, D_{\text{audio}}) \in \{(45,5),\ (41.5,8.5),\ (37.5,12.5),\ (33,17),\ (25,25)\}$B.
    
\end{itemize}

In the three-way mixtures, we allocate more vision than audio tokens because open-source vision datasets are typically much larger than audio ones (e.g., Mint-1T~\citep{awadalla2024mint} vs.\ USP~\citep{USP}), so realistic multimodal mixtures are likely vision-skewed.

\textbf{Joint Scaling Law Fitting}. \label{ssec:fitting} To robustly fit the coefficients in Equation~\ref{eq:unified_law}, we jointly optimize all modality-specific scaling exponents ($\alpha_m$, $\beta_m$), cross-modal transfer coefficients \(v_{k\to m}\), and repetition penalties. We minimize the Huber loss \citep{huber1992robust} on log validation loss with the L-BFGS algorithm \citep{nocedal1980updating}. 

%% file: sections/unimodalscalinglaws.tex
\section{Unimodal Scaling Laws}
\label{sec:unimodalscalinglaws}

We first examine the per-modality scaling behaviors, which lets us ask a basic resource-allocation question: for text, vision, and audio, should additional compute be spent on parameters, on tokens, or on scaling both in tandem?

%Although all coefficients are estimated from the full experiment grid, the modality-specific exponents $(\alpha_m,\beta_m)$ determine how each target modality trades off model size and target-modality data.

\textbf{Different modalities reward different compute allocations: text and vision benefit roughly equally from extra parameters and tokens, whereas audio benefits more from parameters.} Table~\ref{tab:scaling_coefficients} reports the irreducible loss $E_m$, the fitted scaling exponents $(\alpha_m, \beta_m)$, and the induced compute-optimal exponents ($a = {\beta_m}/{(\alpha_m + \beta_m)}, b = {\alpha_m}/({\alpha_m + \beta_m})$), assuming $C \propto N\times D$, $N_{\text{opt}} \propto C^a$ and $D_{\text{opt}} \propto C^b$). Vision has the highest scaling exponents in our setup, with $\alpha_{\text{vis}}=0.50$ and $\beta_{\text{vis}}=0.51$ --- both about $50\%$ above the corresponding text values ($\alpha_{\text{text}}=\beta_{\text{text}}=0.34$) and the audio model exponent ($\alpha_{\text{aud}}=0.32$). In other words, increasing either model size or training tokens reduces vision loss faster than for the other two modalities. Moreover, as $\alpha_{\text{vis}}$ and $\beta_{\text{vis}}$ are close, vision's compute-optimal split is balanced ($a=0.50, b=0.50$). Audio has a more asymmetric profile ($\alpha_{\text{aud}}=0.32$, $\beta_{\text{aud}}=0.52$): the data term shrinks faster than the model term, and the induced compute-optimal split tilts toward model size ($N_{\text{opt}}\!\propto\!C^{0.62}$, $D_{\text{opt}}\!\propto\!C^{0.38}$). Text has $\alpha=\beta=0.34$, giving a balanced compute-optimal split ($a=b=0.50$) close to the original Chinchilla finding \citep{hoffmann2022trainingcomputeoptimallargelanguage}; therefore, standard text-only recipes remain reasonable for the text portion of a multimodal mixture. Relative to existing scaling laws, while text and vision compute-optimality exponents look very similar, our audio exponents sit in a more model-heavy regime, plausibly reflecting (i) the architectural shift from dense to sparse MoE (compared to SODA \cite{manakul2026scaling}) and (ii) higher per-token information density in our high-quality, filtered corpora compared to web-scale data \citep{idahl2026propella, subramanyam2025scaling}.

\begin{table}[h]
\centering
\caption{Fitted scaling-law exponents for each modality in comparison to prior art: Chinchilla \citep{hoffmann2022trainingcomputeoptimallargelanguage} and TGL \citep{tian2025towards} for text, NMM \citep{shukor2025scaling} for vision, and SODA \citep{manakul2026scaling} for audio. We report the irreducible loss $E_m$, the scaling exponents $\alpha_m,\beta_m$, and the induced compute-optimal exponents $a,b$.}
\label{tab:scaling_coefficients}
\begin{tabular}{lccc|cc}
\toprule
\textbf{Modality / Reference} & $E_m$ & $\alpha_m$ & $\beta_m$ & $N_{\text{opt}} \propto C^a$ & $D_{\text{opt}} \propto C^b$ \\
\midrule
Text   & $1.91$ & $0.34$  & $0.34$ & $0.50$  & $0.50$ \\
Vision & $5.67$ & $0.50$  & $0.51$ & $0.50$  & $0.50$ \\
Audio  & $4.36$ & $0.32$  & $0.52$ & $0.62$  & $0.38$ \\
\midrule
Chinchilla \cite{hoffmann2022trainingcomputeoptimallargelanguage} (text-dense) & $1.69$ & $0.34$ & $0.28$ & $0.46$ & $0.54$ \\
TGL \cite{tian2025towards} (text-MoE) & -- & -- & -- & $0.51$ & $0.49$ \\
NMM \cite{shukor2025scaling} (vision-dense) & $1.90$ & $0.30$ & $0.34$ & $0.53$ & $0.47$ \\
SODA \cite{manakul2026scaling} (audio-dense) & -- & -- & -- & $0.37$ & $0.58$ \\
\bottomrule
\end{tabular}
\end{table}

%% file: sections/results.tex
\section{Modality Data Mixing: Transfer, Interference, and Repetition}
\label{sec:results}

After examining how each modality scales with parameters and data, we subsequently investigate the effects of the presence of non-target modalities. We primarily seek to answer the following three practical mixture-design questions: (1) \textit{Which modalities help or hurt one another?} (2) \textit{How quickly repeated non-text data saturates?} (3) \textit{Which modalities are most sensitive to sharing model capacity?}

\begin{figure}[htbp]
    \centering
    \includegraphics[width=0.97\textwidth]{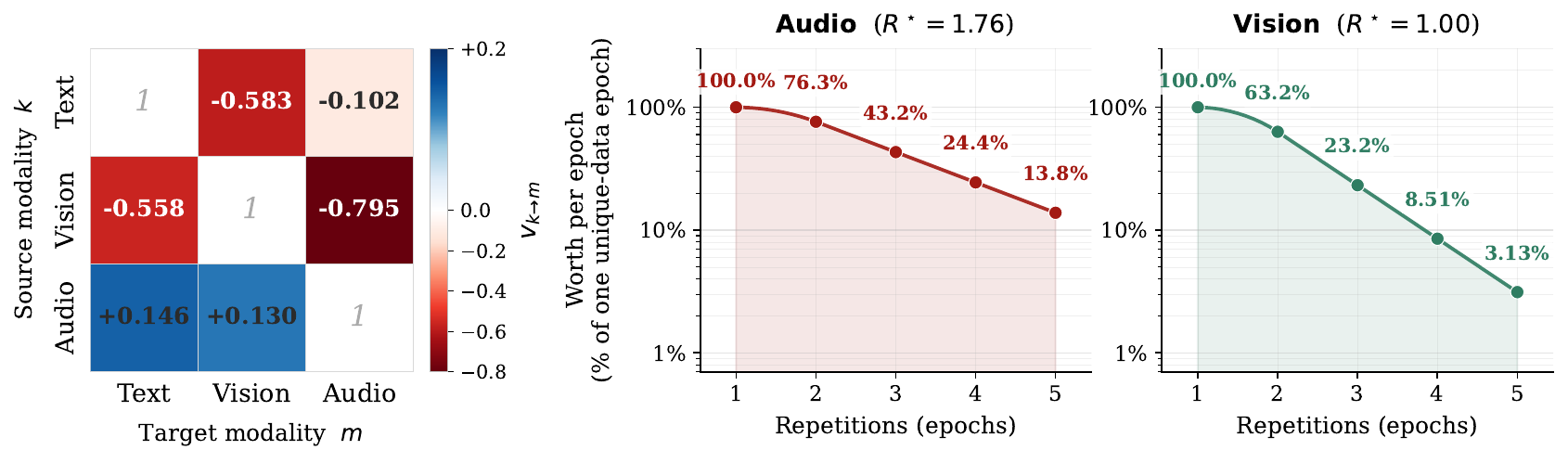}
    \caption{\textbf{(Left)} Cross-Modal Transfer Matrix of directional log-transfer coefficients \(v_{k\to m}\); positive values (blue) indicate synergistic transfer and negative values (red) indicate interference. \textbf{(Middle, Right)} Marginal worth of each successive epoch for audio and vision, expressed as a fraction of one unique-data epoch. Repetition beyond two epochs yields steeply diminishing returns: by the third epoch, each new pass is worth $\le\!43.2\%$ unique data for audio and $\le\!23.2\%$ unique data for vision.}
    \label{fig:lambda}
\end{figure}

\textbf{Cross-modal transfer is asymmetric: audio yields no significant transfer to text and vision, while vision is the most disruptive source, particularly for audio.}  Figure~\ref{fig:lambda} (Left) presents the transfer coefficients $v_{k\to m}$ from our unified scaling law. {The joint fit assigns audio a small positive transfer to both text ($v_{audio\to text} = 0.146$) and vision ($v_{audio\to vision} = 0.130$), but neither coefficient is stable across leave-one-scale-out folds as their standard deviations exceed their means ($v_{audio\to text} = 0.009 \pm 0.37$, $v_{audio\to vision} = 0.104 \pm 0.23$; Table~\ref{tab:f4_interp_cv}). We therefore view audio as an inert modality for the other two modalities.} In contrast, text and vision exhibit mutual negative interference, with $v_{vision\to text} = -0.558$ and $v_{text\to vision} = -0.583$.\footnote{{Interestingly, refitting the vision-side terms ($A,B,E,v_{\text{text}\to\text{vision}},v_{\text{audio}\to\text{vision}}$) separately to image understanding (text-token loss on image--caption pairs) and image generation (image-token loss on caption--image pairs) yields $v_{\text{text}\to\text{vision}}$$=+0.400/-0.716$ and $v_{\text{audio}\to\text{vision}}=+0.346/+0.052$ (understanding/generation). Thus, both text and audio transfer positively to image understanding, and the negative joint text$\to$vision transfer is driven primarily by image generation.}} This means, for a mixture containing 1T text tokens and 100B vision tokens, the 100B vision tokens effectively cancel the effect of 50.72B text tokens ($=-0.558\times(1T/(1T+100B))\times100B$, Section \ref{ssec:transfer}). Vision tokens induce an even more pronounced penalty on audio performance ($v_{vision\to audio} = -0.795$), whereas the interference by text tokens is relatively mild ($v_{text\to audio} = -0.102$). Practically, this makes vision the most expensive modality to mix in when audio and text are also evaluation targets.

% \hspace{-1em}

\textbf{Repeating non-text data beyond two epochs is rarely worth the compute}. Our scaling laws give $R_{\text{vision}}^\star=1.00$ and $R_{\text{audio}}^\star=1.76$. Since $\phi_m(r_m)\to 1+R_m^\star$ as $r_m\to\infty$, this caps the value of repetition at about $2.0$ fresh-data-equivalent epochs for vision and $2.76$ for audio, which means repeating cannot substitute for more than roughly one extra epoch's worth of fresh vision tokens or two of fresh audio tokens. Figure~\ref{fig:lambda} (middle, right) makes this concrete by showing the worth of each successive epoch as a fraction of one unique-data epoch. For vision, the second epoch contributes $63\%$ of a fresh epoch, the third $23\%$, and the fifth only $3\%$; audio decays more gradually, with the third epoch worth $43\%$ and the fifth $14\%$. The same pattern is visible in Figure~\ref{fig:vision_audio_rep}: the $5\text{B}\times 5$ ($25$B tokens seen) curves track below or at most match the $15$B unique-token curves across the entire model-scale range, despite consuming $67\%$ more sampled tokens. While these findings may seem in contrast to text, where up to four repetitions of data has previously been shown to suffer minimal degradation \cite{muennighoff2025scalingdataconstrainedlanguagemodels}, recent work \citep{jha2026data} finds that MoEs overfit more rapidly than dense models even when repeating text data, suggesting that the low repetition utility we observe may partly reflect model sparsity rather than being inherent to vision and audio. \textit{Overall, beyond two epochs, the freed compute is therefore better spent on additional unique tokens or on a larger model.}

\textbf{Audio is the modality most affected by capacity dilution, reinforcing the case for spending its compute budget on parameters rather than tokens.} Beyond the data-side transfer captured by $v_{k\to m}$, modalities also compete for finite MoE parameter capacity through the capacity-sharing term in Eq.~\ref{eq:capacity}. Our joint fit gives $c_{\text{text}}=-0.46$, $c_{\text{vision}}=-0.85$, and $c_{\text{audio}}=-1.87$. All three are negative, so non-target modality data always shrinks the effective parameter budget available to a target modality.\footnote{{We analyze the expert-routing patterns of our models in Appendix~\ref{app:routing} to provide a mechanistic perspective on this observation. We find no modality-wise expert segregation, suggesting that the dilution arises from modalities sharing the same experts.}} The magnitudes, however, differ markedly. To make these numbers concrete, consider a mixture in which the target accounts for $10\%$ of training tokens ($1-h_m=0.9$). The effective parameter count for the target modality is $N_{\text{eff}}^{(m)} = N\exp(c_m \cdot 0.9)$, giving $\approx 0.66\,N$ for text, $\approx 0.47\,N$ for vision, and only $\approx 0.19\,N$ for audio. With $|c_{\text{audio}}|$ roughly $4\times$ larger than text and $2\times$ larger than vision, \textit{audio's effective capacity is eroded most by the presence of other modalities} --- consistent with its model-heavy, compute-optimal frontier in Table~\ref{tab:scaling_coefficients} ($a_{\text{audio}}=0.62$). Both observations point in the same direction: audio benefits more than the other two modalities from additional parameters, whether they are added by scaling the model or recovered by reducing the share of non-audio data in the mixture. Vision sits between the two, while \textit{text is the least capacity-sensitive, retaining most of its nominal capacity even in heavily skewed multimodal mixtures.} This might be another explanation for the modality-bias dynamics towards text described in prior work \citep{zheng2025mllmsdeeplyaffectedmodality}. 

%% file: sections/validation.tex
\section{Validation at Scale and Practical Implications}
\label{sec:validation}

\textbf{Validation}. We evaluate the fitted law on three held-out runs not used for fitting. The first is a 12.6B-A3.8B MoE trained for 200B tokens with a heavily text-skewed mixture of 185B text, 7.5B vision, and 7.5B audio tokens. This tests the fit of our law by extrapolating along three axes: model size, dataset size, and modality proportions. Our next two runs are continual pretraining (CPT) runs --- a regime our fit has not seen --- at the 8.3B-A2.6B scale, both starting from a text-only checkpoint trained on 100B text tokens.  The first CPT run trains $10$B text $+$ $20$B ($10$B $\times 2$) vision $+$ $20$B ($10$B $\times 2$) audio. The second CPT run trains on $40$B text $+$ $5$B vision $+$ $5$B audio. We use the total ($100$B + CPT) text tokens when using our law to predict losses for the CPT runs.

Across the nine modality-level predictions, predicted and observed validation losses are highly correlated (Pearson $r=0.999$). Eight of nine predictions are within $2.6\%$ of the observed loss, and six are within $1.1\%$ (Appendix Table~\ref{tab:validation}). 
{These error rates and high correlation suggest that the fitted law extrapolates well beyond its training distribution across the three held-out settings we test.}
% These error rates and high correlation between predicted and actual validation losses suggest that the law is robust for practitioners to use, even under differing setups, such as training regimes, modality-specific data curation, and much larger compute budgets. %The main failure case is 

\textbf{Example usage.} Let's assume a practitioner wants to train a 120B-A29B MoE model on 15T tokens and $D_{unq}^{text}=15$T, $D_{unq}^{vision}=1$T, and $D_{unq}^{audio}=500$B. Given that all the modalities are equally important to the practitioner, we minimize $(1/|\mathcal{M}|)\sum_{m \in \mathcal{M}}(L_m-E_m)$, i.e., the sum of residuals for all modalities using SLSQP \cite{kraft1988software} and multistart \cite{rinnooy1987stochastic, rinnooy1987stochastic2}. The results suggest using only 9.7T of the text tokens, repeating 1T unique vision tokens 2.6 times, and repeating 500B audio tokens 5.3 times for the optimal MLLM. The example illustrates how scaling laws fitted using our framework can be used to derive context-specific optimal training mixtures. In Section~\ref{sec:limitations}, we discuss why these fitted coefficients should not be treated as universal.
% The example showcases how \rev{a scaling law derived using our methodology} can be used to derive context-specific optimal training mixtures for MLLM developers. \rev{Section \ref{sec:limitations} mentions why our scaling laws should not be used as is.}

%% file: sections/related_works.tex
\section{Related Works}
\label{sec:related_works}

\textbf{Scaling Laws from Text to Multimodal.} Foundational work in scaling establishes predictable power-law relationships between model capacity, compute, and dataset size for language models \cite{kaplan2020scalinglawsneurallanguage,hoffmann2022trainingcomputeoptimallargelanguage}. ~\citet{henighan2020scalinglawsautoregressivegenerative} show that such scaling behavior extends beyond text to autoregressive image, video, and image-text modeling. Recent work extends these laws to sparse architectures: \citet{wan2026holistic} derive holistic MoE scaling laws to map compute budgets to optimal MoE configurations, \citet{tian2025towards} introduce an ``Efficiency Leverage'' metric linking expert activation and compute via power laws. Scaling laws have also been proposed for individual modalities: vision \cite{shukor2025scaling} and audio \cite{manakul2026scaling}.  Beyond empirical fits, \citet{caballero2023brokenneuralscalinglaws} propose Broken Neural Scaling Laws capturing diverse tasks and modalities. Recent works have also presented scaling behaviors in mixed-modal systems, for example, \citet{aghajanyan2023scaling} explored mixed-modal scaling laws unifying the contributions of modalities, deriving a bi-modal scaling law, while \citet{tian2025navil} and \citet{shukor2025scaling} studied the scaling properties of native, early-fusion multimodal models under various data constraints. \citet{sun2024scalinglawhypothesismultimodal} propose extending decoder-based scaling laws to multimodal models via modality-specific tokenization efficiency. \citet{zhang2024explorelimitsomnimodalpretraining} introduce MiCo, demonstrating emergent abilities across 10 modalities. Under data constraints, \citet{muennighoff2025scalingdataconstrainedlanguagemodels} find that up to 4× repetition remains beneficial for text. However, similar systematic studies to identify the limits of repetition and saturation are lacking for non-text modalities.
%In contrast, reuse limits for non-text modalities remain unclear—\citet{manakul2026scaling} study audio scaling without reuse, and \citet{tian2025navil} show vision benefits from more data but do not identify saturation. \citet{shukor2025scalinglawsoptimaldata} also model optimal data mixtures across language, vision pretraining but leave repetition as future work. % Overall, existing work studies modalities or constraints largely in isolation; a unified framework for joint text–vision–audio scaling in sparse MoEs, including data reuse effects, remains missing.

\textbf{Multimodal Interference and Sparse Architectures.}
Joint training across modalities often induces modality bias, with dense models over-relying on text \cite{zheng2025mllmsdeeplyaffectedmodality,wu2025languageoverrulesrevealingtext}. In architectures like Chameleon \cite{chameleonteam2025chameleonmixedmodalearlyfusionfoundation}, this requires disproportionately large text token budgets to maintain performance. Sparse approaches mitigate this: Mixture-of-Transformers \cite{liang2025mixtureoftransformerssparsescalablearchitecture}, MoMa \cite{lin2024momaefficientearlyfusionpretraining}, and Uni-MoE \cite{li2024unimoescalingunifiedmultimodal} route tokens to modality-aware experts, reducing negative transfer and compute. \citet{zhang2025asymoeleveragingmodalasymmetry} exploit modality asymmetry via specialized expert groups (AsyMoE), while \citet{tong2026beyond} highlight vision–language scaling asymmetries in MoEs. MM1 \cite{mckinzie2024mm1methodsanalysis} shows strong sensitivity to data mixtures and tokenization, but lacks a scaling formulation.

%% file: sections/conclusion.tex
\section{Conclusion}
\label{sec:conclusion}

We introduce a {unified scaling-law framework} for native multimodal MoE pretraining that captures how the amount of data, model capacity, data repetition, and cross-modal interactions shape per-modality loss. Our work shows that multimodal scaling is a resource-allocation problem as well, in which modalities differ in how they benefit from data, compete for capacity, tolerate repetition, and affect one another. Our results lead to several broader takeaways. First, compute-optimal allocation is modality-specific: text and vision favor a uniform Chinchilla-style recipe, while audio benefits more from extra parameters than extra tokens. Second, cross-modal transfer is asymmetric: {audio is inert with respect to text and vision, whereas} text and vision interfere negatively with audio and each other. Third, both vision and audio saturate near two fresh-data-equivalent epochs, after which compute is better spent on more unique tokens or a larger model. Finally, audio is roughly $4\times$ more sensitive than text and $2\times$ more sensitive than vision to capacity sharing, consistent with its model-heavy compute-optimal frontier. More broadly, our work has practical implications for frontier MLLM development, where training recipes are often opaque and compute is too expensive for exhaustive trial-and-error search. Our work can also serve as a planning tool by estimating the cost of adding a modality, revealing when scarce data should be repeated versus replaced with model scaling, and exposing which modalities are likely to dominate or degrade the learned representations. Extending this framework to more modalities, tasks, architectures, and data distributions is a step toward more predictable and transparent multimodal training.

%% file: sections/limitation_and_future_work.tex
\section{Limitations and Future Work} \label{sec:limitations}
In Section~\ref{sec:validation}, we show that our derived scaling law extrapolates to a larger from-scratch model and to a continual-pretraining setup. However, we leave validation at substantially larger scales ($\gg\!10$B active parameters, $>\!1$T tokens), extension to additional modalities such as long-form video, and a mechanistic analysis of why audio interferes so little with the other modalities to future work. % We hope this framework provides a principled starting point for designing data mixtures and compute allocations in the next generation of natively multimodal foundation models. 

More broadly, because many design choices were held fixed in the experiments we ran to fit the coefficients of our scaling law, our contribution is best viewed as a framework for fitting context-specific multimodal scaling laws, rather than a universal law for prescribing exact training mixtures. For example, the fitted coefficients likely depend on our choice of tokenizers (i.e., Mistral-Nemo text tokens \citep{mistralnemo}, Emu3.5 image tokens \citep{cui2025emu35nativemultimodalmodels}, WavTokenizer at 40 tokens/s for audio tokens \citep{ji2024wavtokenizer}) and loss objective (next token prediction), and different choices could shift the absolute values of the exponents and coefficients. Moreover, different MoE architectural hyperparameters and degrees of dataset quality could influence the fitted coefficients. However, we expect the qualitative ordering between models or data mixtures to be robust to these variations \citep{subramanyam2025scaling}. Finally, although modeling bidirectional capacity sharing did not consistently improve extrapolation in our experiments (Table~\ref{tab:f4_variants_extrap_M7}), additional training runs could improve the identifiability of these coefficients and, more generally, reduce the coefficient variance observed in Table~\ref{tab:f4_subset_cv}.

%% file: sections/impact_statement.tex
% \FloatBarrier
% \section*{Impact Statement}

%% file: sections/acknowledgments.tex
\section*{Acknowledgements}
We thank the members of the EPFL NLP Lab for their helpful discussions and feedback on our manuscript. We also gratefully acknowledge the support of the Swiss National Science Foundation (No. 215390), the AI2050 program at Schmidt Sciences (Grant \#G-25-69783), Sony Group Corporation, and the Swiss National Supercomputing Center (CSCS) in the form of an infrastructure engineering and development project. This work was also supported as part of the Swiss AI Initiative by a grant from the Swiss National Supercomputing Centre (CSCS) under project IDs a139 and a145 on Alps. Furthermore, this research is with support from Google.org and the Google Cloud Research Credits program for the Gemini Academic Program. This project is also funded by the European Union (ERC, RESPECT-LM, 101222478). Views and opinions expressed are however those of the author(s) only and do not necessarily reflect those of the European Union or the European Research Council. Neither the European Union nor the granting authority can be held responsible for them. Finally, Ayush acknowledges support from the EPFL AI Center PhD Fellowship Program 2026.

%% file: sections/appendix.tex
\clearpage
\appendix
\onecolumn

{\LARGE\bfseries Appendix}\par\vspace{-2mm}

\section{Impact Statement} \label{app:impact_statement}
This work aims to guide the development of future Multimodal LLMs by providing principled scaling laws for data mixing across text, vision, and audio. Its primary positive impact is a potential substantial reduction in redundant compute (and the associated environmental cost) as practitioners can rely on our insights rather than running expensive ablation runs. 

At the same time, we acknowledge potential negative impacts, such as the possibility that improvements to joint vision–audio training could contribute to higher-quality deepfakes and voice clones, with downstream implications for misinformation and fraud. Our contribution is nonetheless foundational rather than directly capability-unlocking, and we believe the value of openly sharing these insights outweighs the risks of concentrating such knowledge within a handful of frontier labs.

\section{Relation to Additive Bimodal Scaling Laws}
\label{app:proof_bimodal}

We now show that Equation~\ref{eq:unified_law} recovers the structure of prior
bimodal interaction laws under a small-interaction approximation, while
generalizing them to directional, unbalanced, and repetition-aware mixtures.

Consider two modalities \(i\) and \(j\), no repetition, and
\(h_i+h_j=1\). For target modality \(i\),
\[
    L_i
    =
    E_i
    +
    A_iN^{-\alpha_i}
    \exp[-\alpha_i c_i h_j]
    +
    B_iD_i^{-\beta_i}
    \exp[-\beta_i v_{j\to i}h_j].
\]
For small \(|c_i|\) and \(|v_{j\to i}|\),
\[
    L_i
    \approx
    L_i^{(0)}
    -
    \alpha_i c_i h_j
    \frac{A_i}{N^{\alpha_i}}
    -
    \beta_i v_{j\to i}h_j
    \frac{B_i}{D_i^{\beta_i}},
    \label{eq:target_expansion}
\]
where
\[
    L_i^{(0)}
    =
    E_i
    +
    \frac{A_i}{N^{\alpha_i}}
    +
    \frac{B_i}{D_i^{\beta_i}}
\]
is the unimodal scaling law for modality \(i\). The same expansion holds for
target modality \(j\). If we evaluate the mixture loss
\[
    L_{ij}=h_iL_i+h_jL_j,
\]
then
\[
\begin{aligned}
    L_{ij}
    \approx\;&
    h_i L_i^{(0)}
    +
    h_j L_j^{(0)}
    \\
    &-
    h_i h_j
    \left[
        \alpha_i c_i \frac{A_i}{N^{\alpha_i}}
        +
        \alpha_j c_j \frac{A_j}{N^{\alpha_j}}
    \right]
    \\
    &-
    h_i h_j
    \left[
        \beta_i v_{j\to i}\frac{B_i}{D_i^{\beta_i}}
        +
        \beta_j v_{i\to j}\frac{B_j}{D_j^{\beta_j}}
    \right].
\end{aligned}
\]
For balanced mixtures \(h_i=h_j=1/2\), \(D_i=D_j=D/2\), and approximately
matched exponents, this becomes
% \[
\begin{equation} \label{eq:bimodal_reduction}
     L_{ij}
    \approx
    \frac{L_i^{(0)}+L_j^{(0)}}{2}
    +
    \frac{A_{ij}^{\mathrm{cap}}}{N^{\bar{\alpha}}}
    +
    \frac{B_{ij}^{\mathrm{tr}}}{D^{\bar{\beta}}},
\end{equation}

% \]
where \(A_{ij}^{\mathrm{cap}}\) and \(B_{ij}^{\mathrm{tr}}\) are induced
interaction coefficients determined by
\((c_i,c_j,v_{i\to j},v_{j\to i})\) and the unimodal constants. Equation
\ref{eq:bimodal_reduction} has the same ``average unimodal law plus additive
interaction powers'' structure as the bimodal scaling law of
\citet{aghajanyan2023scaling}.

The present formulation is more general along four axes. First, transfer is
directional: \(v_{i\to j}\) and \(v_{j\to i}\) need not be equal. Second, the
law applies to arbitrary mixture weights \(h_i,h_j\), not only balanced
bimodal mixtures. Third, it accounts for repeated data through
\(\phi_m(r_m)\). Fourth, the exponential parameterization guarantees positive
effective data even under negative transfer. The only component not included
in Equation~\ref{eq:unified_law} is a constant asymptotic synergy term analogous to
\(-C_{ij}\) in some joint bimodal laws. We omit this term because our
validation losses are target-modality losses: with unlimited target-modality
data and capacity, auxiliary modalities should not change the irreducible
target loss \(E_m\). If the evaluation objective is instead the joint
likelihood of paired modalities, a constant mutual-information offset can be
added as an additional term.

\begin{figure}[htbp]
    \centering
    \includegraphics[width=\textwidth]{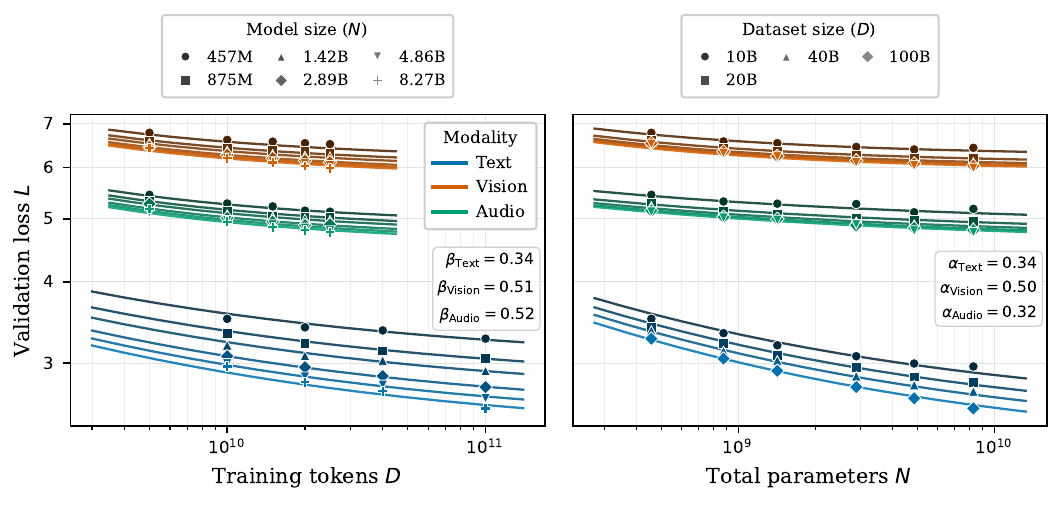}
    \caption{Validation loss as a function of (left) training tokens D at fixed model sizes and (right) total parameter count N at fixed dataset sizes. Text has the lowest irreducible loss $E_m$, while vision has the highest. Vision also exhibits large fitted scaling exponents for both data and model scale, indicating steep improvements with both additional data ($\beta_{vision}$) and model parameters ($\alpha_{vision}$).}
    \label{fig:loss_vs_n_d}
\end{figure}
\section{Learning Rate and Batch Size}
\label{app:lrbs}

\begin{figure}[htbp]
    \centering
    \includegraphics[width=\textwidth]{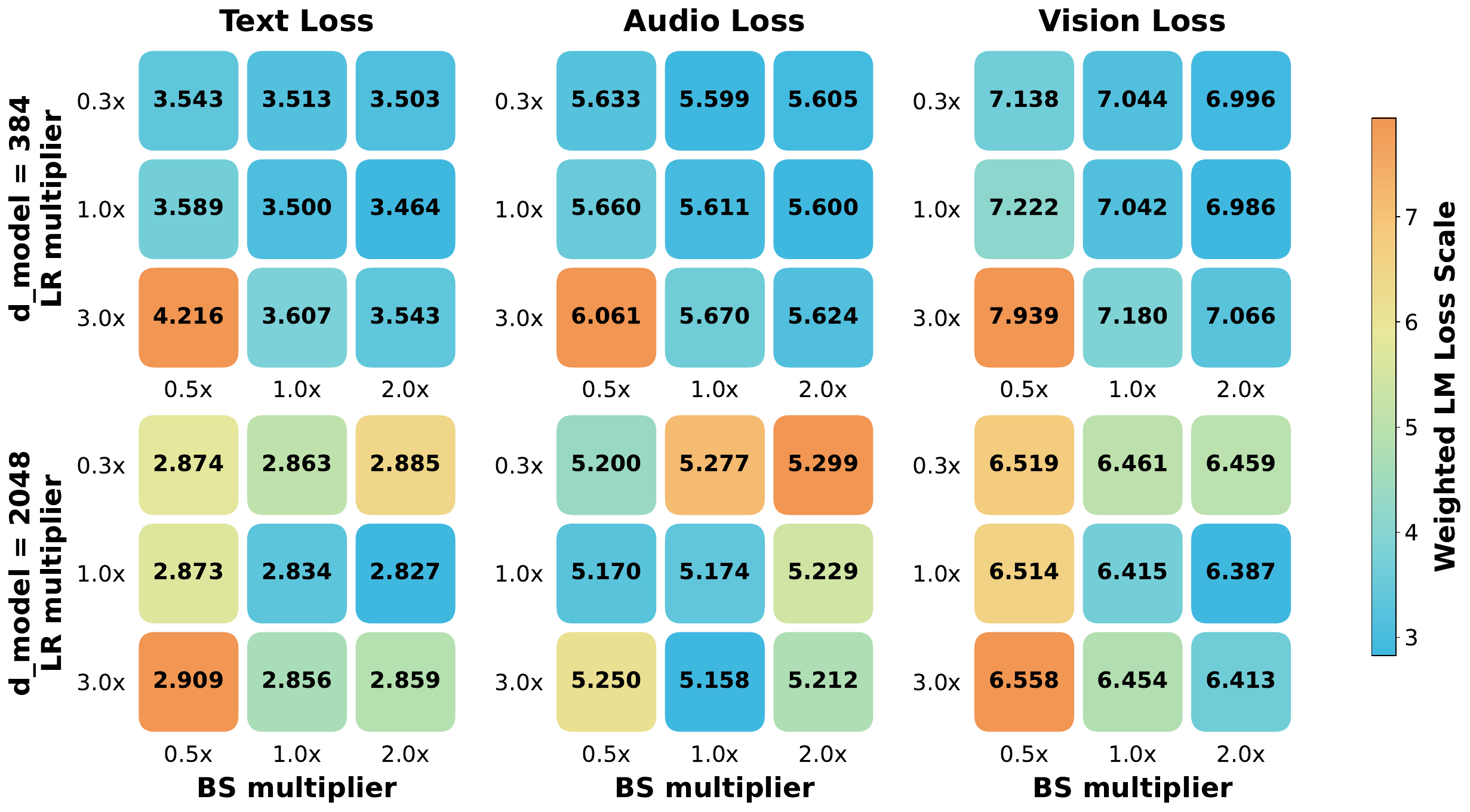}
    \caption{Training losses across different multipliers of optimum learning rates (LR) and batch sizes (BS) obtained using Step Law \citep{li2025predictablescaleistep}. $d_{\text{model}}=384$ and $d_{\text{model}}=2048$ denote the smallest and largest models studied in our work, respectively. The predicted learning rate by Step Law is the best for our case as well, but the optimal batch size is twice the predicted value. It is very close to the best, if not the best, for all three modalities: text, vision, and audio.}
    \label{fig:lrbs}
\end{figure}

We follow the Step Law \citep{li2025predictablescaleistep} to compute the learning rate (LR) and batch size (BS), and treat these computed values as the $1\times$ setting. We then ablate around this point using $\text{LR} \in \{0.3\times, 1\times, 3\times\}$ and $\text{BS} \in \{0.5\times, 1\times, 2\times\}$ (with $1\times$ BS = 4M tokens). We train our smallest (d\_model = 384) and our biggest model (d\_model = 2048) for 30B tokens under each configuration; results are shown in Figure~\ref{fig:lrbs}.

Across both models and all three modalities (text, audio, vision), the combination of $1\times$ LR with $2\times$ BS consistently achieves the lowest or close to the lowest loss. Smaller batch sizes ($0.5\times$) degrade performance, while increasing to $2\times$ yields stable improvements, suggesting benefits from reduced gradient noise at this scale. The $1\times$ BS setting remains competitive but is generally slightly worse than $2\times$.

We also observe that big models are less sensitive to LR and BS variations than smaller ones, though the same optimal region ($1\times$ LR, $2\times$ BS) holds across both model sizes. Overall, these results support using the Step LR-derived learning rate and moderately increasing the batch size to $2\times$.

\section{Training Configurations}
\label{app:train_config}
Table \ref{tab:model-params} shows the exact sizes of the models. Each of our experiments is run on $64$ Nvidia-GH200 GPUs with $96$GB VRAM each, and the full sweep of $268$ runs amounts to $\sim$$300$k GH200-hours of compute. We use the Megatron-LM \cite{shoeybi2019megatron} framework to train our models. In our preliminary experiments, we saw that micro batch sizes (\textsc{MBS}) greater than 1 lead to a non-trivial speedup, and therefore we use expert-parallelism (\textsc{ep}) to achieve \textsc{mbs}$>1$ for all the models. Global batch sizes (\textsc{gbs}) are chosen to be divisible by \textsc{ep}$\times$\textsc{edp} (Expert Data Parallelism), while being close to discovered optimal batch sizes from Appendix~\ref{app:lrbs}. Table~\ref{tab:training-params} contains the training hyperparameters in detail.

\begin{table}[b]
  \centering
  \caption{Hidden size and parameter counts (millions) for all the models used in this study. The active and total parameter count only consists of the non-embedding parameters. Model hyperparameters other than the hidden dimension ($d_{\text{model}}$) are mentioned in Section \ref{sec:experiments}.}
  \label{tab:model-params}
  \begin{tabular}{rrrr}
    \toprule
    $d_{\text{model}}$ & \textbf{Active (M)} & \textbf{Total (M)} & \textbf{Embedding (M)} \\
    \midrule
    $384$  &   $54.5$ &  $252.7$ &  $204.7$ \\
    $576$  &  $122.4$ &  $568.3$ &  $307.0$ \\
    $768$  &  $217.5$ & $1010.2$ &  $409.3$ \\
    $1152$ &  $489.0$ & $2272.7$ &  $614.0$ \\
    $1536$ &  $869.1$ & $4040.0$ &  $818.7$ \\
    $2048$ & $1544.7$ & $7181.8$ & $1091.6$ \\
    \bottomrule
  \end{tabular}
\end{table}

\begin{table}[!h]
  \centering
  \caption{Training hyperparameters for different models. Model labels follow the \textsc{(total params)-A(active params)} convention with both counts including untied embeddings. \textsc{ep}, \textsc{mbs}, and \textsc{gbs} denote the size of the expert parallel group, micro batch size, and the global batch size, respectively. Model hyperparameters other than the hidden dimension ($d_{\text{model}}$) are mentioned in Section \ref{sec:experiments}. }
  \label{tab:training-params}
  \small
  \setlength{\tabcolsep}{5pt}
  \begin{tabular}{llrccrlr}
    \toprule
    \textbf{Model name} & \textbf{Model} & $d_{\text{model}}$ & \textbf{\textsc{ep}} & \textbf{\textsc{mbs}} & \textbf{\textsc{gbs}} & \textbf{Learning Rate} & \textbf{Tokens (B)} \\
    \midrule
    \multirow{8}{*}{M1} & \multirow{8}{*}{457M-A259M} & \multirow{8}{*}{$384$}  & \multirow{8}{*}{$1$} & \multirow{8}{*}{$4$}
        & $256$ & $1.73 \times 10^{-3}$ &   $5$ \\
    & & & & & $256$ & $2.15 \times 10^{-3}$ &  $10$ \\
    & & & & & $256$ & $2.43 \times 10^{-3}$ &  $15$ \\
    & & & & & $256$ & $2.65 \times 10^{-3}$ &  $20$ \\
    & & & & & $256$ & $2.84 \times 10^{-3}$ &  $25$ \\
    & & & & & $256$ & $3.28 \times 10^{-3}$ &  $40$ \\
    & & & & & $512$ & $4.35 \times 10^{-3}$ & $100$ \\
    & & & & & $768$ & $4.93 \times 10^{-3}$ & $150$ \\
    \cmidrule(lr){1-8}
    \multirow{8}{*}{M2} & \multirow{8}{*}{875M-A429M} & \multirow{8}{*}{$576$}  & \multirow{8}{*}{$2$} & \multirow{8}{*}{$2$}
        & $256$ & $9.74 \times 10^{-4}$ &   $5$ \\
    & & & & & $256$ & $1.21 \times 10^{-3}$ &  $10$ \\
    & & & & & $256$ & $1.36 \times 10^{-3}$ &  $15$ \\
    & & & & & $256$ & $1.49 \times 10^{-3}$ &  $20$ \\
    & & & & & $256$ & $1.60 \times 10^{-3}$ &  $25$ \\
    & & & & & $384$ & $1.84 \times 10^{-3}$ &  $40$ \\
    & & & & & $512$ & $2.44 \times 10^{-3}$ & $100$ \\
    & & & & & $768$ & $2.77 \times 10^{-3}$ & $150$ \\
    \cmidrule(lr){1-8}
    \multirow{8}{*}{M3} & \multirow{8}{*}{1.4B-A0.6B} & \multirow{8}{*}{$768$}  & \multirow{8}{*}{$2$} & \multirow{8}{*}{$4$}
        & $256$ & $6.51 \times 10^{-4}$ &   $5$ \\
    & & & & & $256$ & $8.05 \times 10^{-4}$ &  $10$ \\
    & & & & & $256$ & $9.12 \times 10^{-4}$ &  $15$ \\
    & & & & & $256$ & $9.96 \times 10^{-4}$ &  $20$ \\
    & & & & & $256$ & $1.07 \times 10^{-3}$ &  $25$ \\
    & & & & & $256$ & $1.23 \times 10^{-3}$ &  $40$ \\
    & & & & & $512$ & $1.63 \times 10^{-3}$ & $100$ \\
    & & & & & $768$ & $1.85 \times 10^{-3}$ & $150$ \\
    \cmidrule(lr){1-8}
    \multirow{8}{*}{M4} & \multirow{8}{*}{2.9B-A1.1B} & \multirow{8}{*}{$1152$} & \multirow{8}{*}{$2$} & \multirow{8}{*}{$4$}
        & $256$ & $3.63 \times 10^{-4}$ &   $5$ \\
    & & & & & $256$ & $4.49 \times 10^{-4}$ &  $10$ \\
    & & & & & $256$ & $5.08 \times 10^{-4}$ &  $15$ \\
    & & & & & $256$ & $5.55 \times 10^{-4}$ &  $20$ \\
    & & & & & $256$ & $5.95 \times 10^{-4}$ &  $25$ \\
    & & & & & $256$ & $6.87 \times 10^{-4}$ &  $40$ \\
    & & & & & $512$ & $9.10 \times 10^{-4}$ & $100$ \\
    & & & & & $768$ & $1.03 \times 10^{-3}$ & $150$ \\
    \cmidrule(lr){1-8}
    \multirow{8}{*}{M5} & \multirow{8}{*}{4.9B-A1.7B} & \multirow{8}{*}{$1536$} & \multirow{8}{*}{$2$} & \multirow{8}{*}{$3$}
        & $192$ & $2.40 \times 10^{-4}$ &   $5$ \\
    & & & & & $192$ & $2.98 \times 10^{-4}$ &  $10$ \\
    & & & & & $192$ & $3.37 \times 10^{-4}$ &  $15$ \\
    & & & & & $192$ & $3.68 \times 10^{-4}$ &  $20$ \\
    & & & & & $192$ & $3.94 \times 10^{-4}$ &  $25$ \\
    & & & & & $384$ & $4.55 \times 10^{-4}$ &  $40$ \\
    & & & & & $576$ & $6.03 \times 10^{-4}$ & $100$ \\
    & & & & & $768$ & $6.83 \times 10^{-4}$ & $150$ \\
    \cmidrule(lr){1-8}
    \multirow{8}{*}{M6} & \multirow{8}{*}{8.3B-A2.6B} & \multirow{8}{*}{$2048$} & \multirow{8}{*}{$2$} & \multirow{8}{*}{$2$}
        & $256$ & $1.60 \times 10^{-4}$ &   $5$ \\
    & & & & & $256$ & $1.97 \times 10^{-4}$ &  $10$ \\
    & & & & & $256$ & $2.24 \times 10^{-4}$ &  $15$ \\
    & & & & & $256$ & $2.44 \times 10^{-4}$ &  $20$ \\
    & & & & & $256$ & $2.62 \times 10^{-4}$ &  $25$ \\
    & & & & & $384$ & $3.02 \times 10^{-4}$ &  $40$ \\
    & & & & & $512$ & $4.00 \times 10^{-4}$ & $100$ \\
    & & & & & $768$ & $4.53 \times 10^{-4}$ & $150$ \\
    \bottomrule
  \end{tabular}
\end{table}

\section{Datasets}
\label{app:datasets}
We use high-quality permissive datasets for each modality to maximize signal in our limited-scale experiments and to ensure training stability, as lower-quality multimodal data is known to exhibit unstable training dynamics \citep{yao2025efficient}. The data composition for each modality is deliberately aligned with the distribution mixtures used in real-world multimodal large language model (MLLM) pretraining, following the methodologies of Apertus \citep{hernandez2025apertus} for text, Chameleon \citep{chameleonteam2025chameleonmixedmodalearlyfusionfoundation} for vision, and Voxtral \citep{liu2025voxtral} for audio.  Specifically, our text corpus is constructed to balance general knowledge, reasoning, and coding capabilities; it comprises 50\% FineWeb-HQ and 32.5\% FineWeb2-HQ \citep{messmer2025multilingdatacomp} for general text, supplemented by 7.5\% MegaMath-Web-Pro \citep{zhou2025megamath} and 10\% Stack v2-Edu \citep{kandpal2025common, lozhkov2024starcoder}. For the vision modality, we sample from the LLaVA-OneVision-1.5 Mid-Training dataset \citep{an2025llavaonevision15fullyopenframework}, employing an even split of 50\% standard image-caption pairs and 50\% reversed caption-image pairs to encourage bidirectional cross-modal alignment. Finally, our audio training mixture consists of 50\% audio-only samples from the Unsupervised People Speech dataset \citep{USP}. The remaining 50\% uses paired speech--text data in two formats: interleaved audio-text sequences (also inspired by \citep{nguyen-etal-2025-spirit} and \citep{zeng2025scaling}) when the underlying data supports segment- or turn-level composition, and direct ASR-style audio--text pairs otherwise. The interleaved subset includes People's Speech \citep{DBLP:journals/corr/abs-2111-09344}, WenetSpeech \citep{9746682}, LegCo Speech\footnote{\href{https://huggingface.co/datasets/laubonghaudoi/legco-speech}{LegCo Speech}}, CoRal V3 conversation data \citep{coral2024}, and Emilia-YODAS \citep{He_2025}. The direct ASR-style subset includes AISHELL-1\&3 \citep{8384449, shi21c_interspeech}, Multilingual LibriSpeech \citep{pratap20_interspeech}, HUI-Audio-Corpus-German \citep{10.1007/978-3-030-87626-5_15}, Common Voice 24\footnote{\href{https://commonvoice.mozilla.org/en}{Common Voice 24}}, and CoRal V3 read-aloud data \citep{coral2024}. Together, these datasets provide coverage across English, Mandarin, Cantonese, German, other European languages, and broader multilingual speech.

% \section{Specific modality conditioned on another numbers/curves}

\section{Effect of Data Repetition}

\subsection{Effect of repeating vision data}
\begin{figure}[h]
    \centering
    \includegraphics[width=0.99\textwidth]{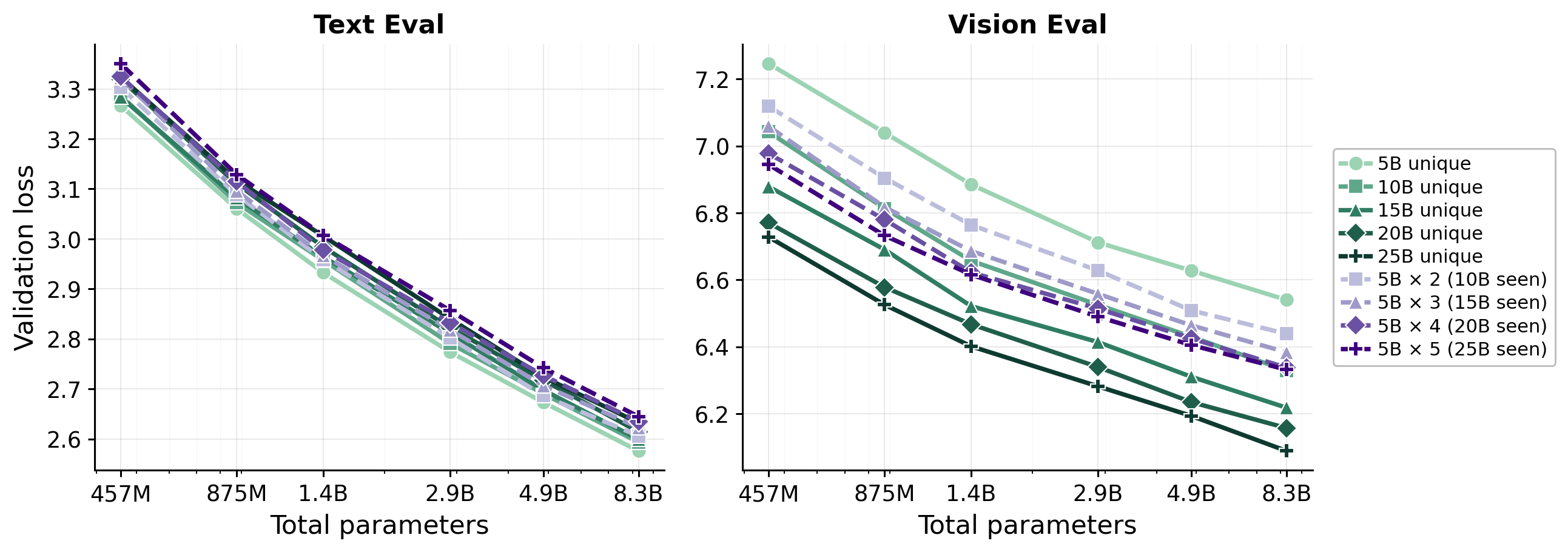}
    \caption{Validation loss as a function of model size for the text+vision repetition experiments at $100$B total training tokens, on the text-only evaluation set (left) and averaged across the two vision evaluation sets (right). Solid green curves vary the unique-vision-token budget from $5$B to $25$B; dashed purple curves repeat a fixed $5$B unique-vision budget $2$--$5$ times. Text loss is essentially independent of the vision repetition factor, while on the averaged vision panel the gap between repeated and unique-data curves widens with model scale, with $5\text{B}\!\times\!5$ ($25$B sampled) falling below the $15$B unique-token curve at the largest scales.}
    \label{fig:vision-repeat-vs-non-repeat}
\end{figure}

The following are our observations from repetition experiments on vision (Figure~\ref{fig:vision-repeat-vs-non-repeat}): 
\begin{itemize}[leftmargin=1.2em,topsep=0pt,itemsep=2pt,parsep=0pt,partopsep=0pt]
    \item Text-only performance remains relatively stable regardless of the composition of the vision data. The variance in text loss across all unique and repeated vision configurations across all modal configurations is close to $4 \times 10^{-4}$, confirming that repeating vision data, rather than having unique vision tokens, does not induce negative transfer or catastrophic forgetting in the text modality within this MoE architecture.
    \item For cross-modal evaluation experiments (Text-to-Vision and Vision-to-Text), increasing unique vision tokens strictly improves performance. However, repetition imposes a severe penalty at larger model scales. For models with parameters more than half a billion, repeating 5B tokens 5 times (25B total) performs worse than training using just 60\% of that token volume (15B) in unique data.
    \item At the smallest model scales (e.g., 457M and 875M total parameters), the performance gap between repeated data (e.g., 5B repeated 2x) and unique data (10B unique) is nearly indistinguishable. This suggests that smaller models are capacity-bound and saturate their representational limits before the penalty of data repetition takes full effect, whereas larger models are strictly data-bound and penalize repetition heavily.
\end{itemize}

\subsection{Effect of repeating audio data}
\begin{figure}[h]
    \centering
    \includegraphics[width=0.99\textwidth]{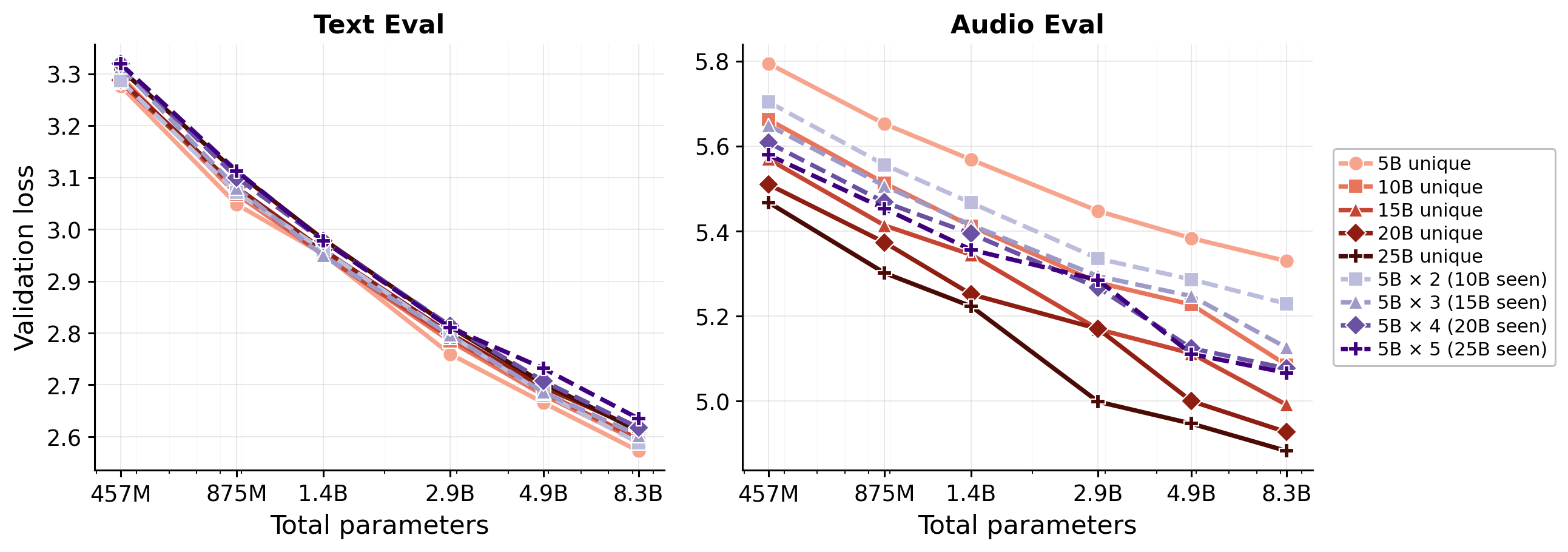}
    \caption{Validation loss as a function of model size for the text+audio repetition experiments at $100$B total training tokens, on the text-only evaluation set (left) and averaged across the audio evaluation sets (right). Solid red curves vary the unique-audio-token budget from $5$B to $25$B; dashed purple curves repeat a fixed $5$B unique-audio budget $2$--$5$ times. As with vision, text loss is essentially unaffected by the audio repetition factor, while on the averaged audio panel repeated data underperforms unique data more strongly at larger model scales: $5\text{B}\!\times\!5$ tracks roughly the $10$B unique-token curve, indicating that $\sim\!60\%$ of the additional compute spent on repetition produces little marginal value.}
    \label{fig:audio-repeat-vs-non-repeat}
\end{figure}

Repeating audio data has similar effects to repeating vision data (Figure~\ref{fig:audio-repeat-vs-non-repeat}). Across all audio evaluations (Audio-only, Audio-Text, and Text-Audio), repeating data yields steeply diminishing returns. Repeating a 5B token dataset 5 times (25B total tokens) achieves an evaluation loss roughly equivalent to training using just 40\% of that token volume (10B) in unique data. This indicates that 60\% of the compute spent on this repetition provides almost zero marginal value. By the same token, the repetition penalty observed in audio perfectly mirrors the behavior observed in vision. In both continuous modalities, repetition beyond 2 epochs simply wastes compute.

\subsection{Interference between modalities}
\begin{figure}[h]
    \centering
    \includegraphics[width=0.99\textwidth]{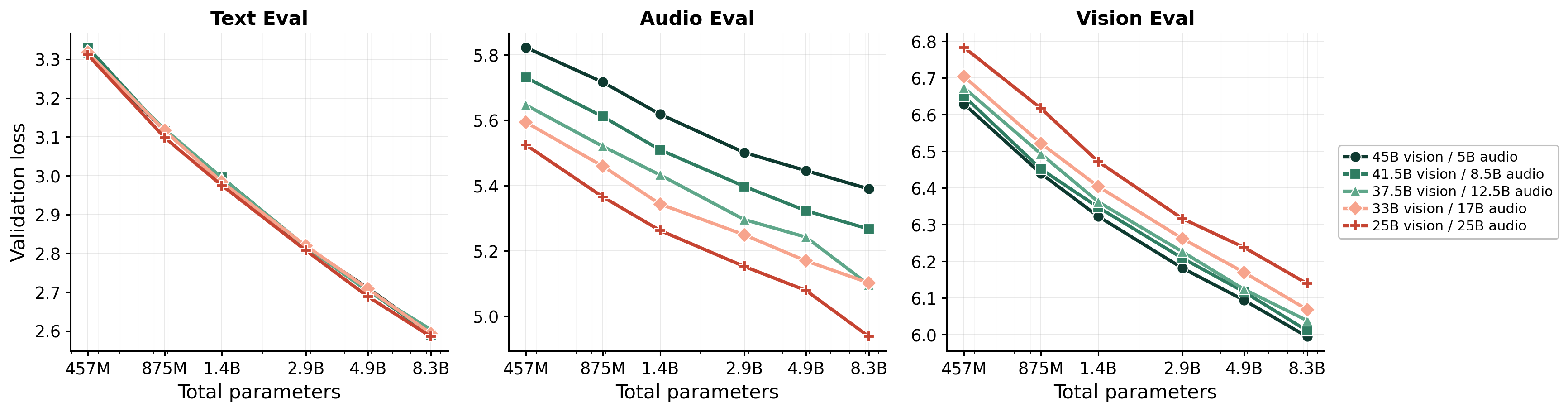}
    \caption{Validation loss as a function of model size for the vision--audio interference experiments under a fixed $150$B-token budget with text held constant at $100$B. The remaining $50$B tokens are traded between vision and audio in five splits ($45$B/$5$B, $41.5$B/$8.5$B, $37.5$B/$12.5$B, $33$B/$17$B, $25$B/$25$B), running from vision-dominant (green) to audio-dominant (red). Panels show loss on the text-only evaluation set (left), averaged across the audio evaluation sets (middle), and averaged across the vision evaluation sets (right).}
    \label{fig:interference}
\end{figure}

Consistent with our repetition experiments, text evaluation remains entirely flat across the varying mixture ratios (Figure~\ref{fig:interference}), proving that text modality seems agnostic to the changing proportions of vision and audio tokens in the remaining budget. For both vision and audio evaluations, performance curves follow a trivial downward trend corresponding directly to the absolute volume of data provided for that modality. The curve is so predictable and linear, suggesting that MoE routing effectively isolates the modalities, mitigating the potential gradient conflicts seen in dense models \cite{liang2024mixture}.

\begin{table}[t]
\centering
\caption{Refits across four interpolation leave-one-out holdouts (M2, M3, M4, M5, with the extremes M1 (smallest) and M6 (largest) always included in training). The \textbf{Joint-fit} column in (a) reports coefficients from the fit on all configurations (M1-M6), while \textbf{Leave-one-scale-out} coefficient estimates are averaged over the four leave-one-out folds and reported as mean $\pm$ standard deviation. The metrics in (b) are computed on runs at the scale held out in each fold.}
\label{tab:f4_interp_cv}
\small
\begin{subtable}{\textwidth}
\centering
\rev{\caption{Fitted coefficients.}}
\begin{tabular}{lcc}
\toprule
\textbf{Coefficient} & \textbf{Joint-fit} & \textbf{Leave-one-scale-out} \\
\midrule

\multicolumn{3}{l}{\textit{Base law (per modality)}} \\
$E_\text{Text}$             & $1.910$  & $1.920 \pm 0.127$ \\
% $A_\text{Text}$           &          & $1362.770 \pm 327.618$ \\
% $B_\text{Text}$           &          & $2835.150 \pm 4286.817$ \\
$\alpha_\text{Text}$        & $0.337$  & $0.359 \pm 0.012$ \\
$\beta_\text{Text}$         & $0.337$  & $0.323 \pm 0.083$ \\

$E_\text{Vision}$           & $5.670$  & $5.727 \pm 0.051$ \\
% $A_\text{Vision}$         &          & $9.666\!\times\!10^{4} \pm 6.053\!\times\!10^{4}$ \\
% $B_\text{Vision}$         &          & $2.562\!\times\!10^{5} \pm 1.592\!\times\!10^{5}$ \\
$\alpha_\text{Vision}$      & $0.568$  & $0.598 \pm 0.037$ \\
$\beta_\text{Vision}$       & $0.508$  & $0.574 \pm 0.047$ \\

$E_\text{Audio}$            & $4.365$  & $4.295 \pm 0.171$ \\
% $A_\text{Audio}$          &          & $906.970 \pm 595.239$ \\
% $B_\text{Audio}$          &          & $8.940\!\times\!10^{5} \pm 1.778\!\times\!10^{6}$ \\
$\alpha_\text{Audio}$       & $0.324$  & $0.359 \pm 0.058$ \\
$\beta_\text{Audio}$        & $0.519$  & $0.467 \pm 0.169$ \\

\multicolumn{3}{l}{\textit{Cross-modal transfer}} \\
$v_{V\!\to\!T}$             & $-0.558$ & $-0.635 \pm 0.245$ \\
$v_{T\!\to\!V}$             & $-0.583$ & $-0.651 \pm 0.088$ \\
$v_{A\!\to\!T}$             & $+0.146$ & $0.009 \pm 0.366$ \\
$v_{T\!\to\!A}$             & $-0.102$ & $-0.411 \pm 0.327$ \\
$v_{A\!\to\!V}$             & $+0.130$ & $0.104 \pm 0.231$ \\
$v_{V\!\to\!A}$             & $-0.795$ & $-0.798 \pm 0.093$ \\

\multicolumn{3}{l}{\textit{Repetition saturation}} \\
$R^\star_V$                 & $0.998$  & $1.138 \pm 0.142$ \\
$R^\star_A$                 & $1.758$  & $1.745 \pm 0.132$ \\

\multicolumn{3}{l}{\textit{Capacity dilution}} \\
$c_\text{Text}$             & $-0.457$ & $-0.439 \pm 0.109$ \\
$c_\text{Vision}$           & $-0.846$ & $-0.767 \pm 0.079$ \\
$c_\text{Audio}$            & $-1.873$ & $-1.471 \pm 0.456$ \\

\bottomrule
\end{tabular}
\end{subtable}
% \end{table}
\vspace{8pt}
\begin{subtable}{\textwidth}
\centering
\rev{\caption{Metrics on the held-out scale.}}
\begin{tabular}{lccccc}
\toprule
\multirow{2}{*}{\textbf{Holdout}} & \multirow{2}{*}{\textbf{RMSE$_\text{log}$}} & \multicolumn{4}{c}{\textbf{Maximum relative error (\%)}} \\
\cmidrule(lr){3-6}
 & & \textbf{Overall} & \textbf{Text} & \textbf{Vision} & \textbf{Audio} \\
\midrule
M2 & $0.0060$ & $2.31$ & $1.25$ & $1.41$ & $2.31$ \\
M3 & $0.0066$ & $1.25$ & $1.25$ & $1.23$ & $1.11$ \\
M4 & $0.0073$ & $2.06$ & $1.45$ & $1.19$ & $2.06$ \\
M5 & $0.0068$ & $1.95$ & $1.77$ & $1.04$ & $1.95$ \\
\bottomrule
\end{tabular}
\end{subtable}
\end{table}

\section{Goodness-of-Fit Results and Scaling Law Variants}
\label{app:validation}

We test the goodness of the fit through (i) leave-one-out cross-validation across model scales (Table~\ref{tab:f4_interp_cv}), {(ii) random-subset resampling of the configuration grid (Table~\ref{tab:f4_subset_cv}),} ({iii}) training extrapolation to the largest scale (Table~\ref{tab:f4_extrap}), ({iv}) ablations of each modeling component (Table~\ref{tab:f4_ablations}), and ({v}) a directional-capacity variant of the law (Tables~\ref{tab:f5_directional} and~\ref{tab:f4_variants_extrap_M7}). Together these establish that the fit is stable across folds, extrapolates to scales beyond its training range, and that each of the three modeling components --- cross-modal transfer, capacity dilution, and repetition saturation --- contributes to predictive accuracy.

We report two goodness-of-fit metrics throughout this section. $\mathrm{RMSE}_{\log}$ is the root-mean-squared error between predicted and observed log loss, $\mathrm{RMSE}_{\log}=\big(\tfrac{1}{n}\sum_{i=1}^{n}(\log \hat{L}_i-\log L_i)^2\big)^{1/2}$, and summarises the fit over all $n$ evaluated runs. The maximum relative error, $\max_{i}|\hat{L}_i-L_i|/L_i$, instead reports the single worst-predicted run in the evaluated set, and is the metric we use when we quote a worst-case error in the main text.

% ==== Table: random-subset resampling ====

\begin{table}[t]
% \color{red}
\centering
\caption{Coefficients under random-subset resampling: five independent refits, each on a random $90\%$ subset of the fitted configurations. The \textbf{Joint-fit} column reports coefficients fit on the full data. \textbf{Random-subset Refit} estimates are averaged over five $90\%$ subsamples of runs and reported as mean $\pm$ standard deviation. Coefficients that contain zero within one standard deviation of the average value are marked$^{\dagger}$.}
\label{tab:f4_subset_cv}
\small
\begin{tabular}{lcc}
\toprule
\textbf{Coefficient} & \textbf{Joint-fit} & {\textbf{Random-subset Refits}} \\
\midrule
\multicolumn{3}{l}{\textit{Base law (per modality)}} \\
$E_\text{Text}$ & $1.910$ & $2.010 \pm 0.085$ \\
$\alpha_\text{Text}$ & $0.337$ & $0.360 \pm 0.048$ \\
$\beta_\text{Text}$ & $0.337$ & $0.401 \pm 0.062$ \\
$E_\text{Vision}$ & $5.670$ & $5.675 \pm 0.068$ \\
$\alpha_\text{Vision}$ & $0.568$ & $0.535 \pm 0.067$ \\
$\beta_\text{Vision}$ & $0.508$ & $0.561 \pm 0.049$ \\
$E_\text{Audio}$ & $4.365$ & $4.331 \pm 0.203$ \\
$\alpha_\text{Audio}$ & $0.324$ & $0.332 \pm 0.066$ \\
$\beta_\text{Audio}$ & $0.519$ & $0.524 \pm 0.173$ \\
\multicolumn{3}{l}{\textit{Cross-modal transfer}} \\
$v_{V\!\to\!T}$ & $-0.558$ & $-0.595 \pm 0.104$ \\
$v_{T\!\to\!V}$ & $-0.583$ & $-0.605 \pm 0.088$ \\
$v_{A\!\to\!T}$ & $+0.146$ & $+0.043 \pm 0.292^{\dagger}$ \\
$v_{T\!\to\!A}$ & $-0.102$ & $-0.240 \pm 0.118$ \\
$v_{A\!\to\!V}$ & $+0.130$ & $-0.014 \pm 0.206^{\dagger}$ \\
$v_{V\!\to\!A}$ & $-0.795$ & $-0.825 \pm 0.174$ \\
\multicolumn{3}{l}{\textit{Repetition saturation}} \\
$R^\star_V$ & $0.998$ & $1.086 \pm 0.164$ \\
$R^\star_A$ & $1.758$ & $1.589 \pm 0.174$ \\
\multicolumn{3}{l}{\textit{Capacity dilution}} \\
$c_\text{Text}$ & $-0.457$ & $-0.479 \pm 0.125$ \\
$c_\text{Vision}$ & $-0.846$ & $-0.839 \pm 0.117$ \\
$c_\text{Audio}$ & $-1.873$ & $-1.737 \pm 0.187$ \\
\bottomrule
\end{tabular}
\end{table}

\paragraph{Cross-validation (Interpolation) across model scales.}
We perform a four-fold leave-one-out cross-validation over the four middle scales (M2, M3, M4, M5), keeping the smallest (M1) and largest (M6) in the training set in every fold so the held-out scale is always interpolated. Table~\ref{tab:f4_interp_cv} reports the fitted coefficients (mean$\pm$std across folds) and per-fold validation metrics on the held-out scale. The base-law and capacity-dilution coefficients are stable across folds (e.g., $\alpha_\text{Text}=0.359\pm0.012$, $E_\text{Text}=1.92\pm0.13$, $c_\text{Text}=-0.439\pm0.11$), and the held-out validation loss is matched to within $2.31\%$ in every fold. The fit therefore does not depend on any single intermediate scale and is stable to interpolation.

{\paragraph{Random-subset resampling of the configuration grid.}
We also test the stability of the law by refitting it on five independent random subsets containing $90\%$ of the fitted points, drawn without regard to scale, mixture, or repetition setting. Table~\ref{tab:f4_subset_cv} compares the coefficients estimated from the joint fit across all runs against the average coefficient values estimated from these five subsets. The base-law exponents, capacity-dilution coefficients, and repetition-saturation points are recovered consistently, and the text--vision transfer coefficients retain both their sign and approximate magnitude. The two audio-source transfer coefficients behave differently: $v_{A\!\to\!T}$ and $v_{A\!\to\!V}$ have standard deviations that exceed their means, and $v_{A\!\to\!V}$ changes sign relative to the joint fit.}

\begin{table}[t]
\centering
\caption{Refits under training extrapolation regimes. {Columns \textbf{R\{\textit{x}-1\} (train M1-\textit{x})} denote the regime where runs corresponding to models M1, M2, ... M\textit{x} are used for fitting, whereas the \textbf{Joint-fit} column reports coefficients from the fit on all configurations (M1-M6). The metrics in (b) are validated using M6 runs only (the most distant out-of-sample scale), so the four regimes are directly comparable. Both RMSE$_\text{log}$ and Overall Maximum Relative Error decrease monotonically as more model scales are included in the fit.}}
\label{tab:f4_extrap}
\small
\begin{subtable}{\textwidth}
\centering
\rev{\caption{Fitted coefficients.}}
\begin{tabular}{lccccc}
\toprule
 & \textbf{Joint-fit} & \textbf{R1 (train M1-2)} & \textbf{R2 (train M1-3)} & \textbf{R3 (train M1-4)} & \textbf{R4 (train M1-5)} \\
\midrule
\multicolumn{6}{l}{\textit{Base law (per modality)}} \\
$E_\text{Text}$ & $1.910$ & $2.197$ & $2.179$ & $2.162$ & $1.930$ \\
$\alpha_\text{Text}$ & $0.337$ & $0.438$ & $0.383$ & $0.450$ & $0.317$ \\
$\beta_\text{Text}$ & $0.337$ & $0.305$ & $0.398$ & $0.324$ & $0.367$ \\
$E_\text{Vision}$ & $5.670$ & $5.502$ & $6.066$ & $5.997$ & $5.732$ \\
$\alpha_\text{Vision}$ & $0.568$ & $0.334$ & $1.100$ & $0.830$ & $0.557$ \\
$\beta_\text{Vision}$ & $0.508$ & $0.741$ & $0.668$ & $0.828$ & $0.585$ \\
$E_\text{Audio}$ & $4.365$ & $5.055$ & $5.037$ & $4.862$ & $4.255$ \\
$\alpha_\text{Audio}$ & $0.324$ & $4.252$ & $1.415$ & $0.570$ & $0.281$ \\
$\beta_\text{Audio}$ & $0.519$ & $0.594$ & $0.839$ & $0.936$ & $0.455$ \\
\multicolumn{6}{l}{\textit{Cross-modal transfer}} \\
$v_{V\!\to\!T}$ & $-0.558$ & $-0.037$ & $-0.568$ & $-0.584$ & $-0.277$ \\
$v_{T\!\to\!V}$ & $-0.583$ & $-0.291$ & $-0.899$ & $-0.966$ & $-0.686$ \\
$v_{A\!\to\!T}$ & $0.146$ & $0.750$ & $0.307$ & $0.173$ & $0.457$ \\
$v_{T\!\to\!A}$ & $-0.102$ & $-1.576$ & $-1.086$ & $-0.334$ & $-0.298$ \\
$v_{A\!\to\!V}$ & $0.130$ & $0.017$ & $0.102$ & $0.167$ & $0.051$ \\
$v_{V\!\to\!A}$ & $-0.795$ & $-0.992$ & $-0.728$ & $-0.534$ & $-0.527$ \\
\multicolumn{6}{l}{\textit{Repetition saturation}} \\
$R^\star_V$ & $0.998$ & $1.233$ & $1.197$ & $1.180$ & $1.265$ \\
$R^\star_A$ & $1.758$ & $1.643$ & $1.505$ & $1.274$ & $1.640$ \\
\multicolumn{6}{l}{\textit{Capacity dilution}} \\
$c_\text{Text}$ & $-0.457$ & $-0.662$ & $-0.538$ & $-0.418$ & $-0.706$ \\
$c_\text{Vision}$ & $-0.846$ & $-1.165$ & $-0.844$ & $-0.634$ & $-0.847$ \\
$c_\text{Audio}$ & $-1.873$ & $-0.806$ & $-1.337$ & $-2.113$ & $-1.724$ \\
\bottomrule
\end{tabular}
\end{subtable}

\vspace{8pt}
\begin{subtable}{\textwidth} 
\centering
\caption{Metrics on the held-out M6 scale. $n_\text{train}$ denotes the number of training points in the respective regime.} \label{app:subtable_error_extrap}
\begin{tabular}{lcccccc}
\toprule
\multirow{2}{*}{\textbf{Estimation Regime}} & \multirow{2}{*}{\textbf{$n_\text{train}$}} & \multirow{2}{*}{\textbf{RMSE$_\text{log}$}} & \multicolumn{4}{c}{\textbf{Maximum Relative Error (\%)}} \\
\cmidrule(lr){4-7}
 & & & \textbf{Overall} & \textbf{Text} & \textbf{Vision} & \textbf{Audio} \\
\midrule
R1 (train M1-2) & $80$ & $0.0430$ & $8.27$ & $4.87$ & {$1.19$} & $8.27$ \\
R2 (train M1-3) & $120$ & $0.0349$ & $6.91$ & $3.70$ & $4.03$ & $6.91$ \\
R3 (train M1-4) & $160$ & $0.0217$ & $4.06$ & $3.49$ & $3.96$ & $4.06$ \\
R4 (train M1-5) & $200$ & $0.0094$ & $2.46$ & $2.46$ & $1.78$ & $2.04$ \\
\bottomrule
\end{tabular}
\end{subtable}
\end{table}

\paragraph{Extrapolation to the largest scale.}
We test how far the law extrapolates beyond its fitting range by training on prefixes of the model-size axis and predicting the largest scale (M6). Four regimes (R1--R4 in Table~\ref{tab:f4_extrap}) progressively grow the training prefix from M1--M2 (R1, $80$ runs) up to M1--M5 (R4, $200$ runs); validation is always on M6 so the four regimes are directly comparable. The maximum relative error on M6 decreases monotonically with the training span: $8.27\%$ in R1, $6.91\%$ in R2, $4.06\%$ in R3, and $2.46\%$ in R4. %Coefficient estimates in the most extreme prefix regimes (R1, R2) inflate (e.g., $A_\text{Audio}\sim 10^{34}$ in R1) because the parameter-side term $A/N^{\alpha}$ is weakly identified when only two model scales contribute, but the predictions on M6 remain bounded since this term is small at large $N$. By R4 the coefficients converge to those of the full fit and the law extrapolates within $2.5\%$ of the observed M6 validation loss.

\begin{table}[t]
\centering
\caption{Component ablations of the joint fit on the full training set. The ``$\Delta$ vs Full'' column shows the relative increase in overall RMSE$_\text{log}$ when the corresponding component is removed. {$k$ denotes the parameters to fit in the resulting scaling law. Per-modality RMSE$_\text{log}$ and the overall maximum relative error, both computed on the fitted set, are shown for context. The full law achieves the minimum in the Maximum Relative Error (\textbf{Max. rel. err.}) column.}}
\label{tab:f4_ablations}
\small
\setlength{\tabcolsep}{4pt}
\resizebox{\textwidth}{!}{%
\begin{tabular}{lcccccccc}
\toprule
\textbf{Variant} & $k$ & \textbf{RMSE$_\text{log}$} & $\Delta$ vs \textbf{Full} & \textbf{RMSE$_\text{Text}$} & \textbf{RMSE$_\text{Vision}$} & \textbf{RMSE$_\text{Audio}$} & \textbf{{Max rel.\ err.\ (\%)}} \\
\midrule
Full law & $26$ & $0.0064$ & -- & $0.0066$ & $0.0053$ & $0.0073$ & $2.16$ \\
No repetition & $24$ & $0.0096$ & $+49.4\%$ & $0.0062$ & $0.0112$ & $0.0107$ & $3.62$ \\
No transfer & $20$ & $0.0064$ & $-0.7\%$ & $0.0061$ & $0.0061$ & $0.0070$ & $2.61$ \\
No capacity & $23$ & $0.0089$ & $+38.4\%$ & $0.0077$ & $0.0074$ & $0.0111$ & $5.24$ \\
\bottomrule
\end{tabular}%
}
\end{table}

% ==== Table 4: directional capacity ====

\begin{table}[!h]
\centering
\caption{RMSE$_\text{log}$ and maximum relative errors (\%) for the directional-capacity variant of our scaling law: a separate $c_{k\to m}$ for each (source, target) pair instead of one $c_m$ per target in Eq. \ref{eq:capacity}. We use the same training extrapolation regimes as Table~\ref{tab:f4_extrap} where \textbf{R\{\textit{x}-1\} (train M1-\textit{x})} denote the regime where runs corresponding to models M1, M2, ... M\textit{x} are used for fitting, and the metrics are computed on M6 only. $n_\text{train}$ denotes the number of training points in the respective regime. {Even though the RMSE$_\text{log}$ and overall maximum relative errors are comparable, the vision errors are strictly worse than the non-directional version (Table~\ref {app:subtable_error_extrap}).}}
\label{tab:f5_directional}
\small

\centering
\begin{tabular}{lcccccc}
\toprule
\multirow{2}{*}{\textbf{Regime}} & \multirow{2}{*}{\textbf{$n_\text{train}$}} & \multirow{2}{*}{\textbf{RMSE$_\text{log}$}} & \multicolumn{4}{c}{\textbf{Maximum relative error (\%)}} \\
\cmidrule(lr){4-7}
 & & & {\textbf{Overall}} & {\textbf{Text}} & {\textbf{Vision}} & {\textbf{Audio}} \\
\midrule
R1 (train M1-2) & $80$ & $0.0441$ & $7.91$ & $5.41$ & $3.88$ & $7.91$ \\
R2 (train M1-3) & $120$ & $0.0358$ & $6.54$ & $3.89$ & $5.60$ & $6.54$ \\
R3 (train M1-4) & $160$ & $0.0208$ & $4.84$ & $3.35$ & $4.84$ & $3.62$ \\
R4 (train M1-5) & $200$ & $0.0101$ & $2.54$ & $2.54$ & $2.21$ & $2.13$ \\
\bottomrule
\end{tabular}
% \end{subtable}
\end{table}

\begin{table}[b]
\centering
\caption{{Metrics on the held-out M6 scale} across the four training extrapolation regimes (R1: train M1-2, R2: train M1-3, R3: train M1-4, R4: train M1-5). Five variants of our law are compared: the full law, three variants with each modeling component individually removed, and the directional-capacity variant. \emph{Lower is better in every column.} Under these extrapolation regimes, we observe that the full law and the directional capacity variant perform consistently well across the regimes.}
\label{tab:f4_variants_extrap_M7}
\small
\setlength{\tabcolsep}{4pt}
\begin{tabular}{l|cccc|cccc}
\toprule
& \multicolumn{4}{c|}{\textbf{RMSE$_\text{log}$}} & \multicolumn{4}{c}{\textbf{Maximum relative error (\%)}} \\
\textbf{Variant} & \textbf{R1} & \textbf{R2} & \textbf{R3} & \textbf{R4} & \textbf{R1} & \textbf{R2} & \textbf{R3} & \textbf{R4} \\
\midrule
Full law & $0.0430$ & $0.0349$ & $0.0217$ & $0.0094$ & $8.27$ & $6.91$ & $4.06$ & $2.46$ \\
No transfer & $0.0469$ & $0.0379$ & $0.0138$ & $0.0095$ & $10.55$ & $7.94$ & $3.71$ & $2.22$ \\
No capacity & $0.0436$ & $0.0355$ & $0.0212$ & $0.0139$ & $9.96$ & $7.76$ & $5.58$ & $5.38$ \\
No repetition & $0.0456$ & $0.0362$ & $0.0218$ & $0.0127$ & $9.30$ & $6.74$ & $4.28$ & $2.70$ \\
Directional capacity & $0.0441$ & $0.0358$ & $0.0208$ & $0.0101$ & $7.91$ & $6.54$ & $4.84$ & $2.54$ \\
\bottomrule
\end{tabular}
\end{table}

\paragraph{Component ablations.}
Table~\ref{tab:f4_ablations} ablates each of the three modeling components --- cross-modal transfer, capacity dilution, and repetition saturation --- by refitting the law with that component removed on the full training set. Removing repetition saturation increases the overall RMSE$_\text{log}$ by $49.4\%$ and capacity dilution by $38.4\%$; removing cross-modal transfer has little effect on average RMSE$_\text{log}$ but worsens Huber loss and worst-case error from $2.16\%$ to $2.61\%$. Repetition saturation and capacity dilution are therefore critical for fit quality, while cross-modal transfer reduces worst-case error further, though not as much as capacity dilution and repetition saturation.

Table~\ref{tab:f4_variants_extrap_M7} also compares the above law variants across extrapolation regimes (along with a directional-capacity variant discussed in the next section). Leaving out any component performs comparably across regimes R2, R3, and R4, but the full law is much better (8.27\% vs 10.55\%, 9.96\%, 9.30\%) in R1 when only M1 and M2 runs are used to fit the laws. 

\paragraph{Directional-capacity variant.}
Our default formulation models capacity dilution with one coefficient per target modality, $c_m$ ($3$ coefficients in total for the three modalities). A natural generalization is to allow the dilution to depend on which source modality is competing: $N_\text{eff}^{(m)} = N \exp\!\left(\sum_{k\ne m} c_{k\to m} h_k\right)$, with one coefficient per ordered (source, target) pair ($6$ coefficients in total). Tables~\ref{tab:f5_directional} and~\ref{tab:f4_variants_extrap_M7} compare this directional variant against the default and the three component-ablations across the same four extrapolation regimes. The directional variant performs comparably to the default on R1--R3, but slightly worse on R4 ({$2.54\%$ vs $2.46\%$ maximum relative error on M6}), so the additional three coefficients do not improve extrapolation. {Moreover, the vision errors are strictly worse than the non-directional version (Table~\ref {app:subtable_error_extrap}).} We therefore retain the simpler 3-coefficient form throughout the paper.

\section{Scaling Law Validation Results} 
\label{app:validation_at_scale}

\begin{figure}[t]
    \centering
    \includegraphics[width=0.92\textwidth]{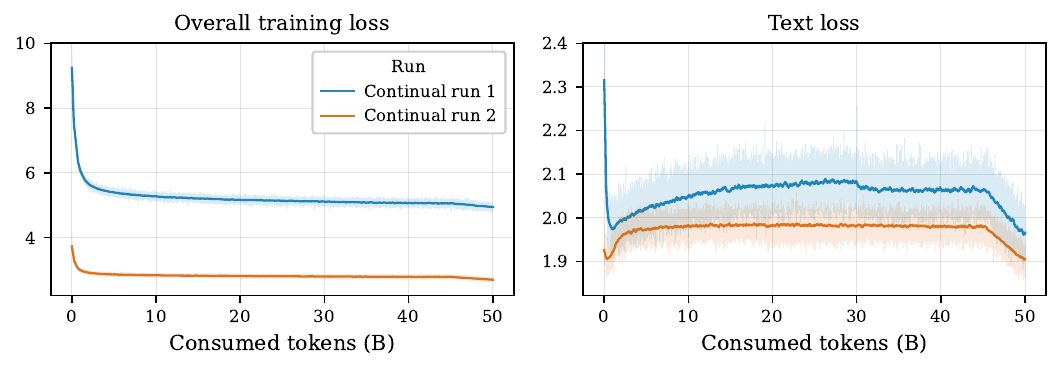}
    \caption{The overall training loss and the text-only loss for (i) \textit{Continual run 1}: Continual pretraining (CPT) on a text-only checkpoint trained with 100B tokens. The continual training mixture consists of 10B text, 20B vision (10B repeated twice), and 20B audio (10B repeated twice), (ii) \textit{Continual Run 2}: Continual pretraining on the same checkpoint as (i) with 40B text, 5B vision, and 5B audio.}
    \label{fig:continual_explanation}
\end{figure}

Table \ref{tab:validation} shows the results on out-of-domain datapoints. All the obtained loss values are very close ($<2.6\%$ error) to their predicted values, barring the test loss in non-text-heavy CPT ($6.73\%$ error). As visible in Figure \ref{fig:continual_explanation}, the text loss increases sharply for the CPT runs, but the non-text-heavy run does not get enough text tokens for the loss to stabilize, unlike the text-heavy CPT run. We hypothesize that longer training would stabilize the text loss and bring it closer to our law’s prediction.

\begin{table}[t]
\centering
\caption{Predicted vs. observed validation loss on three held-out
runs. Predictions use the fitted coefficients from
Table~\ref{tab:scaling_coefficients} and Section~\ref{sec:results}. No parameters are re-tuned on these runs, and 8/9 losses are predicted to within 2.51\%.}

\label{tab:validation}
\small
\begin{tabular}{lcccr}
\toprule
\textbf{Modality} & \textbf{Observed} & \textbf{Predicted} & $\Delta$ & \textbf{Relative error} \\
\midrule
\multicolumn{5}{l}{\textit{12.6B-A3.8B, from scratch on 200B tokens (185B text / 7.5B vision / 7.5B audio)}} \\
Text   & $2.4264$ & $2.4874$ & $+0.0610$ & $+2.51\%$ \\
Vision & $6.4240$ & $6.4185$ & $-0.0055$ & $-0.09\%$ \\
Audio  & $5.2589$ & $5.1470$ & $-0.1119$ & $-2.13\%$ \\
\midrule
\multicolumn{5}{l}{\textit{8.3B-A2.6B, non-text-heavy continual pretraining: 10B text + 20B vision ($r{=}2$) + 20B audio ($r{=}2$)}} \\
Text   & $2.7840$ & $2.5967$ & $-0.1872$ & $-6.73\%$ \\
Vision & $6.2183$ & $6.2249$ & $+0.0067$ & $+0.11\%$ \\
Audio  & $5.0230$ & $5.0178$ & $-0.0053$ & $-0.10\%$ \\
\midrule
\multicolumn{5}{l}{\textit{8.3B-A2.6B, text-heavy continual pretraining: 40B text + 5B vision + 5B audio}} \\
Text   & $2.5897$ & $2.5627$ & $-0.0271$ & $-1.04\%$ \\
Vision & $6.4872$ & $6.5974$ & $+0.1101$ & $+1.70\%$ \\
Audio  & $5.3068$ & $5.3016$ & $-0.0052$ & $-0.10\%$ \\
\bottomrule
\end{tabular}
\end{table}

% \newpage

\section{Routing Analysis}
\label{app:routing}

In Section~\ref{sec:results}, we observed that the capacity dilution coefficients, $c_{\text{text}}=-0.46$, $c_{\text{vision}}=-0.85$, and $c_{\text{audio}}=-1.87$ were all negative, meaning non-target modality data always shrinks the effective parameter budget available to a target modality. To mechanistically investigate capacity dilution, we look at the routing patterns of the models. Following prior work on analysing MoE routing \citep{bandarkar2026multilingualroutingmixtureofexperts,khandelwal2026leveragingroutingdynamicsmixtureofexperts}, we accumulate per-expert assignment counts over each modality-specific validation set, giving a per-layer distribution over experts for every modality, and compare these distributions across modalities using the Jensen--Shannon divergence (JSD), normalised to $[0,1]$.

We observe that routing is diffuse rather than modality-specific. At the M6 scale, the per-layer expert-assignment distributions of the modalities present in the training mixture are nearly indistinguishable: $\mathrm{JSD}(\text{text}, \text{vision}) = 0.003$ and $\mathrm{JSD}(\text{text}, \text{audio}) = 0.003$--$0.004$. Table~\ref{tab:routing_jsd} reports this divergence across model scales, where it stays below $0.03$ everywhere and decreases with model size. We see the following observations hold across the model sizes: (i) every trained modality uses the full set of experts, with the routing distribution remaining close to uniform; and (ii) expert loads are balanced across modalities, with no modality-level imbalance. What little modality separation exists is concentrated in the layers closest to the output and shrinks as model size grows. We hypothesize that the diffuse routing may partly reflect the load-balancing loss. Thus, in our setting, capacity dilution does not appear to arise from modalities partitioning the expert pool, but rather from their sharing of the same experts.

\begin{table}[H]
% \color{red}
\centering
\caption{Mean per-layer Jensen--Shannon divergence between the expert-assignment distributions of two modalities, for every model scale. Mixture labels give the token split between the two modalities present in training. Values near zero indicate that the two modalities route to the same experts in the same proportions; the scale is normalized to $[0,1]$. {Barring the smallest model M1, all the divergences are $\leq0.01$, indicating negligible modality-wise expert segregation. This hints that modalities compete for representational capacity within the same experts leading to $c_m < 0$.}}
\label{tab:routing_jsd}
\small
\setlength{\tabcolsep}{6pt}
\begin{tabular}{lcccc}
\toprule
& \multicolumn{2}{c}{\textbf{Text vs.\ vision}} & \multicolumn{2}{c}{\textbf{Text vs.\ audio}} \\
\cmidrule(lr){2-3} \cmidrule(lr){4-5}
\textbf{Model} & \textbf{95/5} & \textbf{75/25} & \textbf{95/5} & \textbf{75/25} \\
\midrule
M1 & $0.026$ & $0.010$ & $0.014$ & $0.011$ \\
M2 & $0.007$ & $0.005$ & $0.005$ & $0.005$ \\
M3 & $0.010$ & $0.008$ & $0.006$ & $0.009$ \\
M4 & $0.006$ & $0.006$ & $0.007$ & $0.007$ \\
M5 & $0.005$ & $0.005$ & $0.005$ & $0.004$ \\
M6 & $0.003$ & $0.004$ & $0.004$ & $0.003$ \\
\bottomrule
\end{tabular}
\end{table}